%% file: main.tex
\documentclass[10pt,twocolumn,letterpaper]{article}
\PassOptionsToPackage{table}{xcolor}
\usepackage[pagenumbers]{cvpr}
\input{preamble}
\usepackage{balance}
\definecolor{cvprblue}{rgb}{0.21,0.49,0.74}
\usepackage[breaklinks,colorlinks,allcolors=cvprblue]{hyperref}

\title{EviViT: Evidence-Adaptive Vision Transformers for Fine-Grained Perception}
\author{Yaoxin Niu, Zhangquan Chen, Yang Zhang, Xinlei Yu, Xiang An, Zhumei Wang, Chih-Ting Liao, Hongkun Cao, Ruqi Huang}
\hypersetup{
  pdftitle={EviViT: Evidence-Adaptive Vision Transformers for Fine-Grained Perception},
  pdfauthor={Yaoxin Niu, Zhangquan Chen, Yang Zhang, Xinlei Yu, Xiang An, Zhumei Wang, Chih-Ting Liao, Hongkun Cao, Ruqi Huang}
}

\makeatletter
\newcommand{\ps@preprint}{%
  \let\@oddhead\@empty
  \let\@evenhead\@empty
  \def\@oddfoot{\hbox to\textwidth{\rlap{\footnotesize Preprint.}\hfil\thepage\hfil}}%
  \let\@evenfoot\@oddfoot
}
\makeatother

\begin{document}
\onecolumn
\thispagestyle{preprint}
\input{Sec/frontmatter}
\input{Sec/abstract}
\twocolumn
\input{Sec/intro}
\input{Sec/pipeline_figure}
\input{Sec/related}
\input{Sec/method}
\input{Sec/exp}
\input{Sec/conclusion}
\FloatBarrier
\input{Sec/statements}
\balance

{\small
\bibliographystyle{ieeenat_fullname}
\bibliography{references}
}

\clearpage
\nobalance
\appendix
\maketitlesupplementary
\input{Sec/appendix}

\end{document}

%% file: preamble.tex
\input{math_commands}

\usepackage{array}
\usepackage{adjustbox}
\usepackage{multirow}
\usepackage{enumitem}
\usepackage{placeins}

\definecolor{basegrey}{RGB}{242,243,245}
\definecolor{eviblue}{RGB}{25,88,166}
\definecolor{gaingreen}{RGB}{20,116,67}
\definecolor{lossred}{RGB}{183,45,38}
\definecolor{pendinggrey}{RGB}{115,120,128}
\newcommand{\method}{EviViT}
\newcommand{\gain}[1]{{\color{gaingreen}#1}}
\newcommand{\loss}[1]{{\color{lossred}#1}}

\newcommand{\tablebodyfont}{\normalsize}
\newcommand{\tablenotefont}{\normalsize}

%% file: math_commands.tex
\usepackage{amsmath,amsfonts,bm}

\def\eqref#1{equation~\ref{#1}}

\def\1{\bm{1}}

\DeclareMathAlphabet{\mathsfit}{\encodingdefault}{\sfdefault}{m}{sl}
\SetMathAlphabet{\mathsfit}{bold}{\encodingdefault}{\sfdefault}{bx}{n}



%% file: Sec/frontmatter.tex
\noindent\rule{\linewidth}{1.5pt}
\begin{center}
{\fontsize{16}{19}\selectfont\bfseries
EviViT: Evidence-Adaptive Vision Transformers\\
for Fine-Grained Perception\par}
\vspace{0.14in}
\rule{\linewidth}{0.5pt}\par
\vspace{0.15in}
{\normalsize
\textbf{Yaoxin Niu}$^{1,2,\dagger}$ \quad
\textbf{Zhangquan Chen}$^{1,\dagger}$ \quad
\textbf{Yang Zhang}$^{3}$ \quad
\textbf{Xinlei Yu}$^{4}$ \quad
\textbf{Xiang An}$^{5}$\\[0.35em]
\textbf{Zhumei Wang}$^{6}$ \quad
\textbf{Chih-Ting Liao}$^{7}$ \quad
\textbf{Hongkun Cao}$^{2}$ \quad
\textbf{Ruqi Huang}$^{1,*}$\par}
\vspace{0.11in}
{\small
$^{1}$ Tsinghua University \quad $^{2}$ Peng Cheng Laboratory\\
$^{3}$ The Hong Kong University of Science and Technology \quad $^{4}$ Chinese University of Hong Kong\\
$^{5}$ MiniMax \quad $^{6}$ Beijing Institute of Technology \quad $^{7}$ University of New South Wales\\[0.45em]
$^{\dagger}$ Equal contribution. \quad $^{*}$ Corresponding author.\\[0.35em]
Code: \href{https://github.com/YXNiu/EviViT}{\textcolor{eviblue}{github.com/YXNiu/EviViT}}\\[0.2em]
Data: \href{https://huggingface.co/datasets/YXNiu/Human-Search-Traces}{\textcolor{eviblue}{huggingface.co/datasets/YXNiu/Human-Search-Traces}}\par}
\end{center}
\vspace{0.13in}

\noindent\begin{minipage}{\textwidth}
\captionsetup{hypcap=false}
\centering
\includegraphics[width=0.97\linewidth]{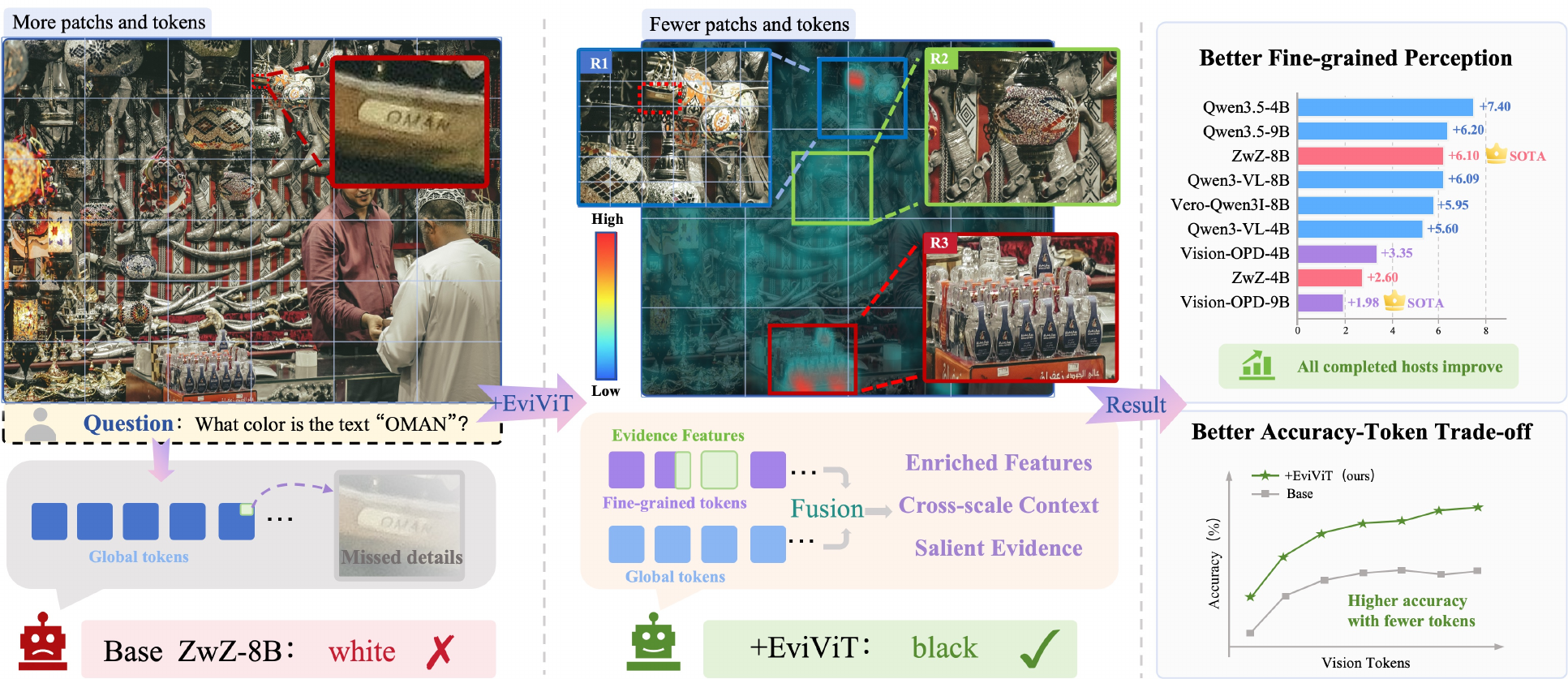}
\captionof{figure}{\textbf{Question-conditioned detail allocation.} A global-only model
misses the small queried text despite processing more patches. EviViT predicts
an evidence density learned from human search annotations, re-reads selected
regions at native detail, and fuses them with the global scene. In the paired budget sweep, both Qwen3-VL models improve accuracy while
using fewer visual tokens on average than their global-only baselines.}
\label{fig:teaser}
\end{minipage}
\par\vspace{0.7em}

%% file: Sec/abstract.tex
\begin{abstract}
Fine-grained visual perception enables vision--language models to distinguish subtle attributes and ground their answers in visual evidence. In high-resolution scenes, raising the resolution of the entire image spends visual tokens on irrelevant regions, while isolated crops can lose the context needed to interpret the selected evidence. We introduce \method{}, a lightweight attachment that learns where a pretrained vision transformer should acquire detail. Human visual-search traces supervise a question-conditioned evidence density, which guides regional re-reading from the original pixels and the allocation of visual tokens. A sparse, coordinate-aware bridge then connects the regional features to the global scene, allowing the model to interpret precise evidence in context. Learned with the model backbone frozen, the attachment serves both the base model and compatible post-trained descendants without refitting. Experiments across nine models show consistent gains in average fine-grained accuracy. On Qwen3-VL-4B and 8B, matched-budget comparisons show that \method{} outperforms global-only processing at every tested token ceiling while using fewer visual tokens on average.
\end{abstract}

%% file: Sec/intro.tex
\section{Introduction}
\label{sec:introduction}

Visual understanding is fundamental to vision--language models (VLMs),
providing perceptual grounding for question answering, multimodal reasoning,
and decision making. In
high-resolution scenes, an answer may hinge on a distant identifier, subtle
attribute, short text, or spatial relation in a small image region. Resizing
the image to a fixed visual-token budget can erase such evidence, whereas
uniformly increasing resolution spends more computation on regions irrelevant
to the question. Fine-grained benchmarks such as V$^{\ast}$Bench and HR-Bench
expose this tension~\citep{wu2024v,wang2025divide}. The question is thus not
simply how many visual tokens a model receives, but where it spends them.

Existing approaches take two main directions. The first acquires detail by
tiling the image~\citep{wang2025divide} or by active search. Methods such as
V$^{\ast}$ and Mini-o3 select and inspect crops over several
steps~\citep{wu2024v,lai2026mini}. When the language model directs the search, each new view requires
another round of inference. The second direction prunes encoded
tokens according to the question, merges similar tokens, or retains both
local detail and global context~\citep{li2026occamtoken,tong2026focus}.
These methods can remove redundant
computation, yet selecting encoded tokens cannot recover detail lost during
initial resizing. Together, these limitations motivate a different question:
can the vision encoder itself learn which original-image regions need
additional resolution before answer generation?

Human visual search provides supervision for this decision. Cursor movements
and zooms record both the regions an annotator inspects and the alternatives
they leave behind before settling on a final box. We turn the full search
trace into a question-conditioned evidence density. The map teaches the encoder
which regions deserve a closer look without requiring it to replay the search.

Three challenges follow from this choice.
\textbf{(C1) Locate the evidence.} A human search includes detours as well as
useful clues. The encoder must learn which locations matter for the question.
\textbf{(C2) Choose complementary views.} Zooming in on the same detail twice
spends tokens that could reveal another clue or the context around the first.
\textbf{(C3) Relate detail to the scene.} A tight crop can make small text
legible while hiding the object it labels. The encoder must connect the new
detail to its location in the full image.

To meet these challenges, we introduce EviViT, a lightweight attachment
that teaches a pretrained vision transformer (ViT) which regions of the
original image to re-read before the model answers. As Figure~\ref{fig:teaser}
illustrates, both the visual backbone and the large language model (LLM)
remain frozen. The Prompt-conditioned Token--Evidence Aligner (PTEA) first
aligns question representations with intermediate ViT features to predict a
dense evidence distribution supervised by human search traces. The
Evidence Region Planner selects one or two distinct evidence regions and
a region that supplies context. It divides the local token budget among
these views. The frozen encoder then re-reads them from the original
pixels, where fine detail is still available. Finally, we fuse regional and global features with a Sparse Bridge.
The Bridge uses image coordinates to exchange information between tokens
from the same part of the scene. This lets the language model interpret
readable local evidence within the scene, without iterative crop commands.

Separating evidence acquisition from answering makes the attachment
reusable. We train one attachment per base model using human searches
from the VisualProbe training split. Post-trained descendants
that retain the base model's visual interface can reuse these weights
unchanged. At inference, the model needs only the original image and question.

EviViT improves average fine-grained accuracy on all nine base and
post-trained models. Across the seven benchmark scores, Qwen3-VL-4B
gains 5.60 percentage points and Qwen3-VL-8B gains 6.09. We also compare
both sizes with global-only processing at seven token ceilings from 1K to
20K. At every ceiling, selective re-reading improves mean accuracy on
VisualProbe and V$^{\ast}$Bench while using 90--97\% as many visual tokens
on average. In a separate fine-tuning experiment, EviViT retains its advantage
on VisualProbe and ZoomBench after low-rank adaptation (LoRA) of the
language model~\citep{hu2021lora}.

Our contributions are summarized as follows:
\begin{itemize}
    \item \textbf{Learning where to look from human searches.} We use the full
    search trajectory, not just its final box, to supervise
    question-conditioned resolution allocation inside a pretrained ViT.
    \item \textbf{End-to-end perception with feature reuse.} EviViT reuses global features to select and integrate evidence from the original pixels, enabling one answering call without external crop tools.
    \item \textbf{Reusable and efficient fine-grained perception.} A base model
    and its compatible descendants share one attachment without refitting.
    Selective re-reading improves accuracy across token budgets. The language
    model can continue learning with the attachment frozen.
\end{itemize}

%% file: Sec/pipeline_figure.tex
\begin{figure*}[t]
\centering
    \includegraphics[width=0.91\linewidth]{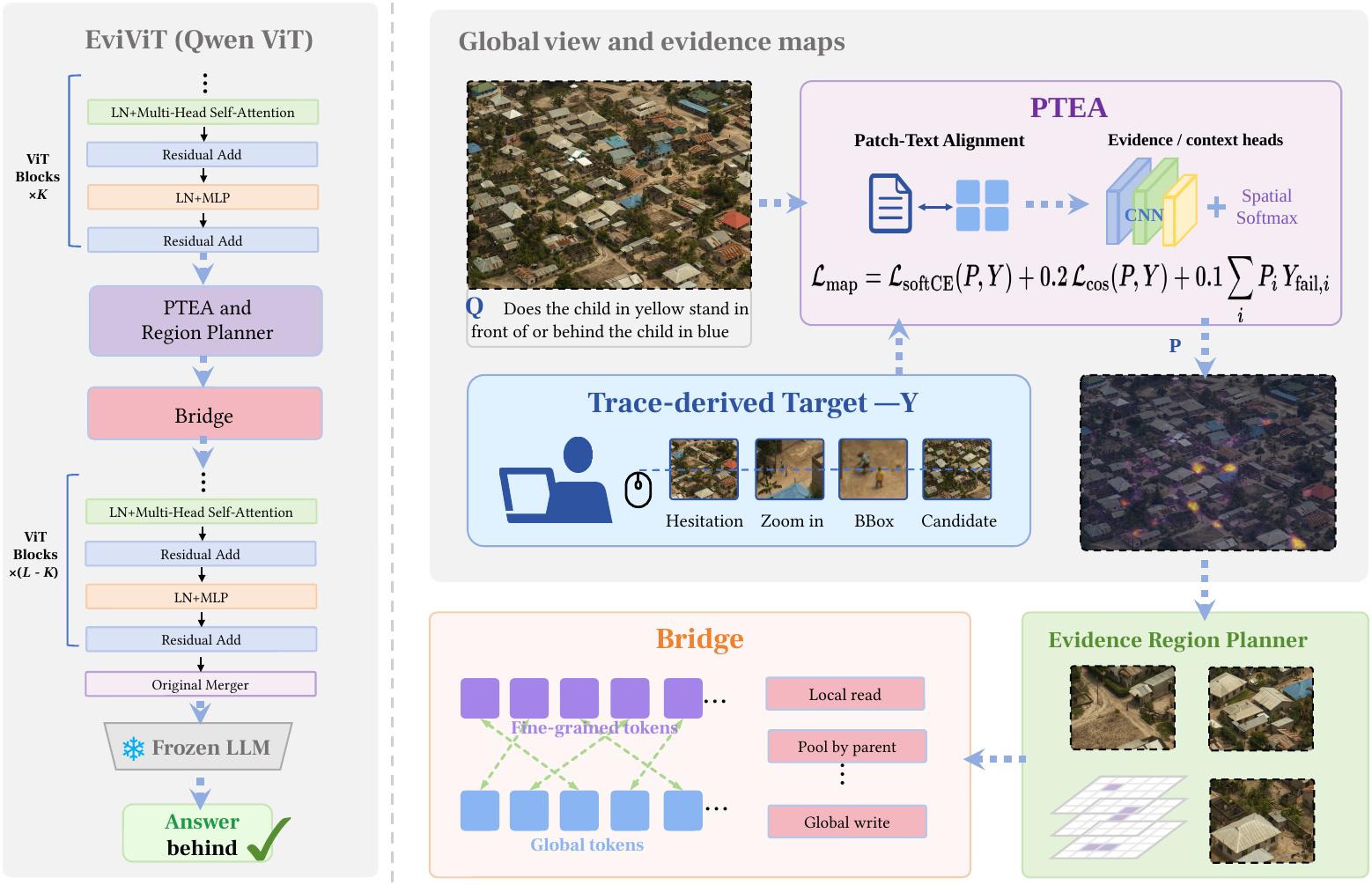}
\caption{\textbf{EviViT pipeline.} Human interactions are distilled into a
pair of spatial targets. PTEA predicts evidence and context densities on the
global feature grid (the heatmap shows the evidence density $P$). The planner chooses one or two decisive regions and one
context region and assigns their input grids. The Bridge exchanges features at
matching source coordinates; the model then completes visual encoding and answers.}
\label{fig:pipeline}
\end{figure*}

%% file: Sec/related.tex
\section{Related work}
\label{sec:related}

\paragraph{Vision encoders.}
Vision Transformers represent images as patch sequences and learn their
interactions through self-attention~\citep{dosovitskiy2020image}. CLIP and SigLIP
learn transferable visual representations from image--text pairs using
contrastive and pairwise sigmoid objectives, respectively
~\citep{radford2021learning,zhai2023sigmoid}. Dynamic-resolution encoders in the
Qwen-VL family accommodate different image sizes with variable-length token
sequences~\citep{wang2024qwen2,bai2025qwen3}. This makes more
detail available, but increasing the resolution of the entire image spends
tokens on both decisive evidence and irrelevant background. EviViT instead
learns where to re-read source pixels for a given question and fuses the
resulting regional tokens with the global grid inside the frozen encoder.

\paragraph{Visual perception.}
High-resolution perception methods recover details through tiling, selective
cropping, or interactive search. V$^{\ast}$ uses language-guided search to
locate small targets~\citep{wu2024v}; FOCUS derives crop relevance from
the model's internal representations~\citep{zhong2026focus}. Mini-o3 scales multi-turn
visual exploration, while AdaptVision learns when to acquire additional crop
tokens through reinforcement learning~\citep{lai2026mini,lin2026adaptvision}.
Other methods use crops during training. Zooming without Zooming (ZwZ)
distills regional supervision into full-image inference, and Vision-OPD uses
a crop-conditioned teacher for on-policy self-distillation
~\citep{wei2026zooming,yuan2026vision}. Search policies request another view while answering, whereas distillation
updates the answering model during training.
EviViT instead learns a visual attachment while keeping that model frozen.
It acquires regional detail before answer generation, without crop commands
from the language model, and retains the global view alongside the local
observations. This separates evidence acquisition from the answering policy,
allowing the same attachment to complement compatible foundation and
post-trained models.

\paragraph{Adaptive attention.}
Adaptive computation changes which visual information is sampled or retained.
DynamicViT prunes tokens using learned importance scores, whereas Token Merging
combines similar tokens~\citep{rao2021dynamicvit,bolya2022token}. Deformable
attention instead learns data-dependent sampling positions
~\citep{xia2022vision}. For VLMs, OccamToken adds query-aware pruning, and
Focus-Scan-Refine combines instruction-relevant local evidence with
complementary global context~\citep{li2026occamtoken,tong2026focus}.
Token compression reduces redundancy, but it can only select or combine
features already available at the input resolution. EviViT also decides which
detail to acquire: its predicted density allocates fresh tokens from the
original pixels while preserving a global context stream.
Learning which regions to revisit calls for spatial supervision. Gaze
datasets provide one source by recording where people look while answering
visual questions~\citep{das2017human,sood2021vqa,chen2021predicting}.
EviViT uses cursor
movements, zoom windows, and final-evidence annotations to supervise a spatial
evidence predictor. The human record therefore guides the allocation of
resolution, rather than prescribing a sequence of actions for the answering
model to imitate.

%% file: Sec/method.tex
\section{Method}
\label{sec:method}

A fine-grained question may hinge on a small object or a few characters of
text. EviViT guides the frozen ViT back to the relevant pixels while keeping
the surrounding scene in view (Figure~\ref{fig:pipeline}). PTEA uses the
question to predict evidence and context maps. The planner chooses which
regions to re-read, and the Bridge fuses their features with those already
computed for the full image.

\subsection{Human visual-search data}
\label{subsec:human_data}

\begin{figure}[!htb]
\centering
\includegraphics[width=\columnwidth]{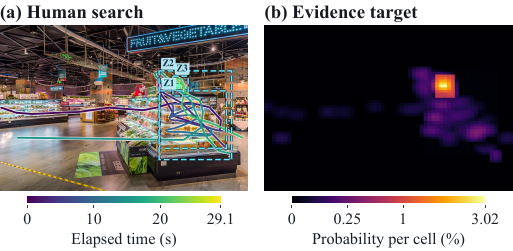}
\caption{\textbf{From a human search to its target.} Left: cursor color marks elapsed time; dashed cyan boxes show zooms and green marks the final box. Right: the training target $Y$ from Eq.~(\ref{eq:trace_map}), shown as probability per grid cell.}
\label{fig:human_trace_h4}
\end{figure}

Figure~\ref{fig:human_trace_h4} shows how an annotator found the price of
a pack of oranges. The cursor path and zooms trace the search across the
shelf, and the final box marks the price label. Team members recorded
1,144 such searches on the VisualProbe training split~\citep{lai2026mini}.
Cursor positions, zooms, resets and boxes use the coordinates of the
original image. Timestamps give the dwell times. At test time, EviViT
needs only the image and question.

\subsection{Learning a spatial evidence density}
\label{subsec:trace_supervision}

\paragraph{From search events to evidence density.}
For an image $I$ and question $q$, we render point events as Gaussian
kernels and boxes as soft regional mass. The spatial maps $Y_k$ summarize the whole search,
the branch after the last reset, the last completed zoom and the final
box. We combine them with fixed weights $w_k$ to form the evidence target
\begin{equation}
  Y=\operatorname{PeakNorm}\!\left(\sum_k w_k Y_k\right).
  \label{eq:trace_map}
\end{equation}
Peak normalization sets the largest value to one. For training and
display, we resize $Y$ to the patch grid and normalize it to unit mass. Abandoned regions yield
a weak negative map $Y_{\rm fail}$.

\paragraph{Predicting where to look.}
A text Transformer contextualizes the question tokens $\hat t_j$.
PTEA projects each global patch feature $v_i$ into the same space with
$W_{\mathrm{img}}$. Cosine similarity and a learned scale $\tau$ determine
the alignment weights
\begin{equation}
  a_{ij}=\operatorname{softmax}_{j}\!\bigl(\tau\cos(W_{\mathrm{img}}v_i,\hat t_j)\bigr).
  \label{eq:patch_text}
\end{equation}
The summary $c_i=\sum_j a_{ij}\hat t_j$ gives more weight to the words
relevant to patch $i$. A convolutional head combines the visual features,
aligned text and coordinates. A spatial softmax turns its scores into the
evidence density $P$. Soft cross-entropy fits $P$ to the normalized target.
A cosine loss with weight $\lambda_1$ rewards spatial agreement, and a
penalty with weight $\lambda_2$ discourages mass on abandoned regions.
The training objective is
\begin{equation}
  \mathcal L_{\rm map}
  =\mathcal L_{\rm CE}+\lambda_1\mathcal L_{\rm cos}
   +\lambda_2\langle P,Y_{\rm fail}\rangle .
  \label{eq:map_loss}
\end{equation}
Supplementary Sec.~\ref{app:attachment_settings} gives the target and
loss weights. We then freeze the shared alignment and evidence branch and
fit a complementary branch on the same features. Its target emphasizes the
broader area the annotator inspected. The evidence map $P$ locates a
likely detail, and the context map $P_{\rm ctx}$ identifies surroundings
that may help interpret it.

\input{Tabs/tab_fine_grained}

\subsection{Evidence Region Planner}
\label{subsec:allocator}

\paragraph{Selecting complementary regions.}
The strongest peaks in $P$ can fall on the same object. The planner
selects the first peak, suppresses its neighborhood and keeps a second
region if it adds distinct evidence. It then masks the selected locations
in $P_{\rm ctx}$ to find a complementary context view. The global view
preserves the layout of all selected regions.

\paragraph{Allocating the local budget.}
One head uses statistics of both maps to reserve a bounded share of the
local budget for context. The remaining allocation favors decisive
regions with greater evidence mass. A second head refines each box using
the question, local features and the proposed box geometry. The encoder's
patch grid determines the final resolution of each view
(Supplementary Sec.~\ref{app:capacity_floor}).

\subsection{Native re-reading and sparse fusion}
\label{subsec:bridge}

EviViT reuses the global features that guided selection and encodes the
new crops up to the same layer. Figure~\ref{fig:regional_features} compares
patch features from visual block~19 of the original ViT and EviViT on a
VisualProbe training image, showing how re-reading captures finer detail
around the clock face.

\par\smallskip\noindent\begin{minipage}{\columnwidth}
\captionsetup{hypcap=false}
\centering
\includegraphics[width=\columnwidth]{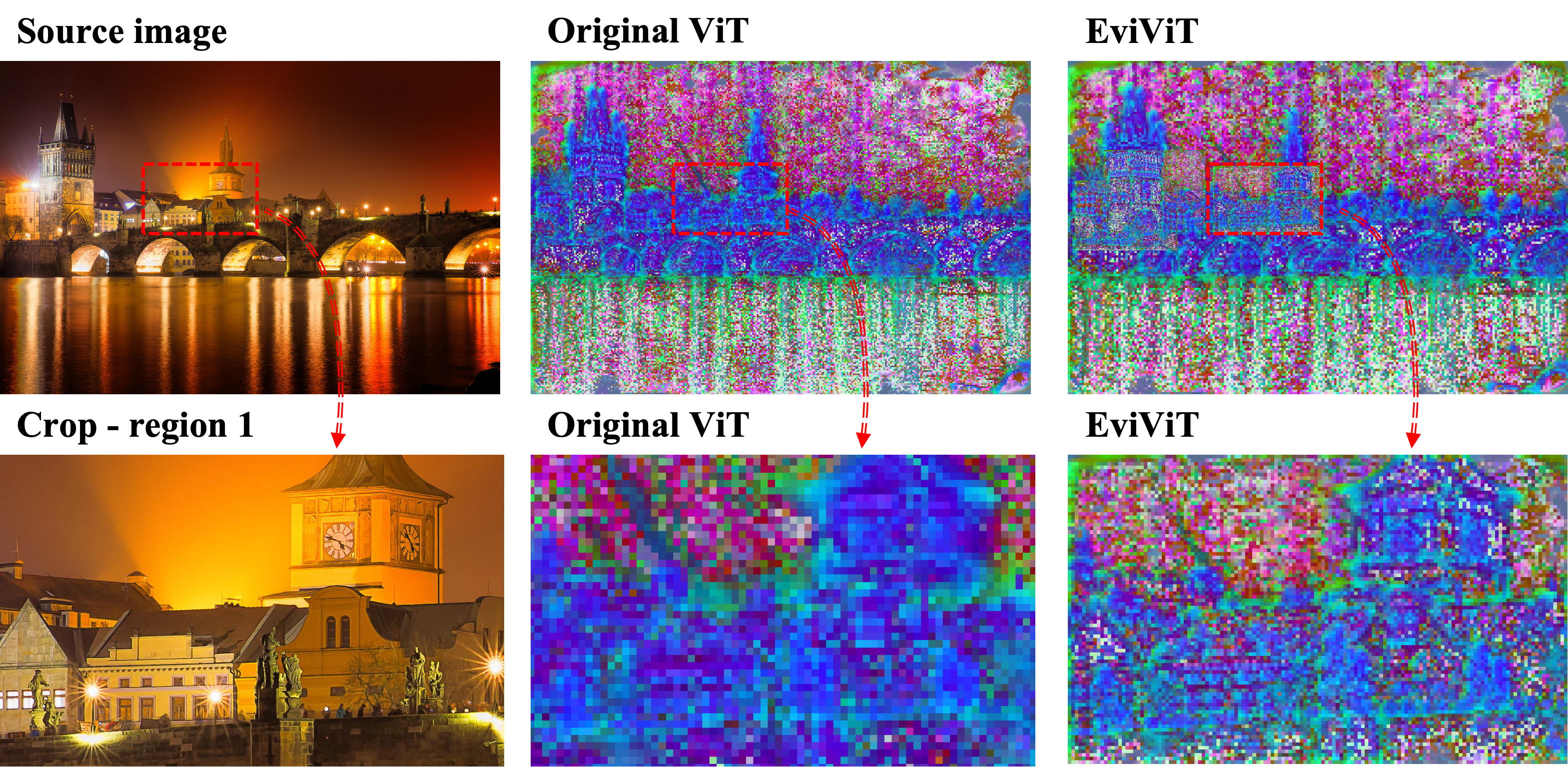}
\captionof{figure}{\textbf{Re-reading a clock tower.} A shared principal component analysis (PCA)
maps both models' visual features to color. Top right: regional features at
source coordinates. Below: enlarged native grid (middle) and re-encoded crop
(right). The colors show feature variation, not attention.}
\label{fig:regional_features}
\end{minipage}\par\smallskip

\input{Sec/strong_model_figure}

To place this detail back in the scene, the Bridge maps each local token
$\ell_i$ to its parent in the global grid. The token attends to a small
neighborhood $\mathcal N_i$ around that parent. For global neighbor $g_j$, the
coordinate bias $\phi(r_{ij})$ encodes relative offset and log scale
$r_{ij}=(\Delta x_{ij},\Delta y_{ij},\Delta s_{ij})$. With $d$-dimensional queries
$q_i=W_q\ell_i$ and keys $k_j=W_k g_j$, the attention weights are
\begin{equation}
  \alpha_{ij}=\operatorname{softmax}_{j\in\mathcal N_i}
  \!\left(\frac{q_i^\top k_j}{\sqrt d}+\phi(r_{ij})\right).
  \label{eq:bridge_weights}
\end{equation}
A local token can now read features from its surroundings, including
those beyond the crop boundary. The Bridge adds the resulting message
to the local feature,
\begin{equation}
  \tilde\ell_i=\ell_i+W_o\sum_{j\in\mathcal N_i}\alpha_{ij}W_v g_j .
  \label{eq:local_read}
\end{equation}
For the reverse update, the Bridge groups local messages by parent and
writes them into the corresponding global positions. Zero-initialized
output projections start the Bridge at identity, and its residual updates
are bounded. Global tokens without local observations keep their values at this step.
Both streams then pass through the remaining frozen blocks. The language model
answers the original question in one call, without a crop tool or an extra turn.

\subsection{Training and portability}
\label{subsec:training_method}

We train PTEA and its context branch on search maps and derive the planner's
budget and box targets from the same traces. Bridge training uses cross-entropy on
reference answers and an identity penalty. The ViT, merger and language model
stay frozen.

We fit one attachment per base model. A post-trained descendant can load
the same weights if it retains the base model's visual blocks, width,
merger and coordinate system. This lets a descendant change how it answers
while keeping what the base attachment learned about where to look.
Supplementary Sec.~\ref{app:frozen_models}
lists the insertion blocks. For a different visual encoder, we use the
same search traces to fit an attachment to its grid
(Supplementary Sec.~\ref{app:cross_encoder}).

%% file: Tabs/tab_fine_grained.tex
\begin{table*}[t]
\centering
\caption{\textbf{Fine-grained perception accuracy (\%).} Gains are in percentage points. Top-three paired gains per benchmark are shaded orange \colorbox{orange!35}{dark}--\colorbox{orange!15}{light}; top-three mean gains are shaded green \colorbox{gaingreen!35}{dark}--\colorbox{gaingreen!15}{light}.}
\label{tab:fine_grained}
\tablebodyfont
\setlength{\tabcolsep}{4.7pt}
\renewcommand{\arraystretch}{1.00}
\begin{tabular}{lccccccccl}
\toprule
Model & VP-E & VP-M & VP-H & V$^{\ast}$ & HR-4K & HR-8K & Zoom & Avg. & {} \\
\midrule
\multicolumn{10}{l}{\textit{Author-reported references; $\ddagger$ cells are matched local API re-evaluations}} \\
GPT-4o$^{\dagger}$ & 47.50 & 15.40 & 11.20 & 71.20$^{\ddagger}$ & 67.12$^{\ddagger}$ & 62.50$^{\ddagger}$ & 47.10$^{\ddagger}$ & 46.00$^{\ddagger}$ & {} \\
Gemini-3-Flash$^{\dagger}$ & 67.38 & 50.75 & 47.17 & 84.82 & 89.25 & 85.50 & 59.41$^{\ddagger}$ & 69.18$^{\ddagger}$ & {} \\
GPT-5.2$^{\dagger}$ & 57.45$^{\ddagger}$ & 38.43$^{\ddagger}$ & 38.68$^{\ddagger}$ & 79.06 & 81.12 & 78.38 & 50.89 & 60.57$^{\ddagger}$ & {} \\
GPT-5.4$^{\dagger}$ & 61.70$^{\ddagger}$ & 29.85$^{\ddagger}$ & 20.75$^{\ddagger}$ & 76.96 & 84.00 & 77.88 & 52.66 & 57.69$^{\ddagger}$ & {} \\
Gemini-3.1-Pro$^{\dagger}$ & 58.16$^{\ddagger}$ & 41.42$^{\ddagger}$ & 41.51$^{\ddagger}$ & 87.96 & 89.63 & 86.88 & 61.18 & 66.68$^{\ddagger}$ & {} \\
Gemini-3.5-Flash$^{\dagger}$ & 63.83$^{\ddagger}$ & 51.87$^{\ddagger}$ & 48.11$^{\ddagger}$ & 89.01 & 89.12 & 86.62 & 61.42 & 70.00$^{\ddagger}$ & {} \\
\midrule
\multicolumn{10}{l}{\textit{Locally evaluated open checkpoints and paired EviViT attachments}} \\
Qwen3-VL-4B & 60.28 & 40.30 & 41.51 & 84.29 & 76.25 & 73.38 & 45.80 & 60.26 &  \\
\rowcolor{basegrey} \quad + EviViT & \textbf{63.83} & \textbf{50.37} & \textbf{48.11} & \cellcolor{orange!15}\textbf{87.96} & \textbf{79.62} & \textbf{77.75} & \cellcolor{orange!15}\textbf{53.37} & \textbf{65.86} & {\color{gaingreen}(+5.60)} \\
Qwen3-VL-8B & 67.38 & 42.16 & 44.34 & 85.86 & 76.00 & 72.75 & 43.91 & 61.77 &  \\
\rowcolor{basegrey} \quad + EviViT & \textbf{69.50} & \cellcolor{orange!25}\textbf{53.36} & \textbf{51.89} & \cellcolor{orange!25}\textbf{91.10} & \cellcolor{orange!15}\textbf{80.25} & \cellcolor{orange!15}\textbf{78.25} & \textbf{50.65} & \textbf{67.86} & {\color{gaingreen}(+6.09)} \\
Qwen3.5-4B & 63.83 & 44.40 & 45.28 & 81.68 & 79.00 & 76.38 & 49.11 & 62.81 &  \\
\rowcolor{basegrey} \quad + EviViT & \cellcolor{orange!25}\textbf{70.92} & \textbf{53.36} & \cellcolor{orange!15}\textbf{55.66} & \cellcolor{orange!35}\textbf{87.96} & \cellcolor{orange!25}\textbf{83.75} & \textbf{81.25} & \cellcolor{orange!35}\textbf{58.58} & \cellcolor{gaingreen!35}\textbf{70.21} & {\color{gaingreen}(+7.40)} \\
Qwen3.5-9B & 65.96 & 44.40 & 42.45 & 84.82 & 76.62 & 73.38 & 51.01 & 62.66 &  \\
\rowcolor{basegrey} \quad + EviViT & \textbf{66.67} & \cellcolor{orange!15}\textbf{54.85} & \textbf{51.89} & \textbf{87.96} & \cellcolor{orange!35}\textbf{82.50} & \cellcolor{orange!35}\textbf{81.12} & \textbf{57.04} & \cellcolor{gaingreen!25}\textbf{68.86} & {\color{gaingreen}(+6.20)} \\
ZwZ-4B & \textbf{69.50} & 47.76 & 34.91 & 90.05 & 78.12 & 76.62 & 55.62 & 64.66 &  \\
\rowcolor{basegrey} \quad + EviViT & 66.67 & \textbf{52.61} & \cellcolor{orange!25}\textbf{46.23} & \textbf{91.10} & \textbf{78.88} & \textbf{77.25} & \textbf{58.11} & \textbf{67.26} & {\color{gaingreen}(+2.60)} \\
ZwZ-8B & 71.63 & 46.64 & 46.23 & 89.53 & 80.50 & 77.12 & 55.98 & 66.80 &  \\
\rowcolor{basegrey} \quad + EviViT & \cellcolor{orange!35}\textbf{79.43} & \textbf{55.97} & \cellcolor{orange!35}\textbf{62.26} & \textbf{92.67} & \textbf{81.12} & \textbf{78.75} & \textbf{60.12} & \cellcolor{gaingreen!15}\textbf{72.90} & {\color{gaingreen}(+6.10)} \\
Vero-Qwen3I-8B & 66.67 & 43.66 & 54.72 & 86.39 & 81.50 & 77.50 & 46.27 & 65.24 &  \\
\rowcolor{basegrey} \quad + EviViT & \textbf{72.34} & \cellcolor{orange!35}\textbf{55.22} & \textbf{61.32} & \textbf{87.43} & \textbf{83.12} & \cellcolor{orange!25}\textbf{83.88} & \cellcolor{orange!25}\textbf{55.03} & \textbf{71.19} & {\color{gaingreen}(+5.95)} \\
Vision-OPD-4B & 73.76 & 49.63 & 51.89 & 86.39 & 81.00 & 76.75 & 59.05 & 68.35 &  \\
\rowcolor{basegrey} \quad + EviViT & \cellcolor{orange!25}\textbf{80.85} & \textbf{52.24} & \textbf{53.77} & \textbf{89.53} & \textbf{82.62} & \textbf{80.75} & \textbf{62.13} & \textbf{71.70} & {\color{gaingreen}(+3.35)} \\
Vision-OPD-9B & \textbf{74.47} & 53.73 & 47.17 & \textbf{93.19} & \textbf{80.88} & 78.62 & 63.79 & 70.26 &  \\
\rowcolor{basegrey} \quad + EviViT & 73.05 & \textbf{58.58} & \textbf{54.72} & 90.05 & 80.62 & \textbf{82.62} & \textbf{66.04} & \textbf{72.24} & {\color{gaingreen}(+1.98)} \\
\bottomrule
\end{tabular}
\vspace{2pt}\parbox{0.98\linewidth}{\tablenotefont $^{\dagger}$Author-reported cells: GPT-4o VisualProbe from Mini-o3~\citep{lai2026mini}; Gemini-3-Flash values from PixelEyes~\citep{gong2026pixeleyes}; GPT-5.2/5.4 and Gemini-3.1-Pro/3.5-Flash values from Vision-OPD~\citep{yuan2026vision}. $^{\ddagger}$Our single-call API evaluations use the local inputs, answer-only prompt, deterministic decoding, and Qwen3-VL-8B judge. Mixed-source averages are descriptive. Local Global uses the native processor (at most $16\times1024^2$ pixels); EviViT's token budget varies with image size (Supplementary Sec.~\ref{app:capacity_floor}). VP-E/M/H denote the Easy, Medium and Hard tiers of VisualProbe. Avg.\ weights all seven scores equally. Zoom uses the full image.}
\end{table*}

%% file: Sec/strong_model_figure.tex
\begin{figure*}[t]
    \centering
    \includegraphics[width=0.95\linewidth]{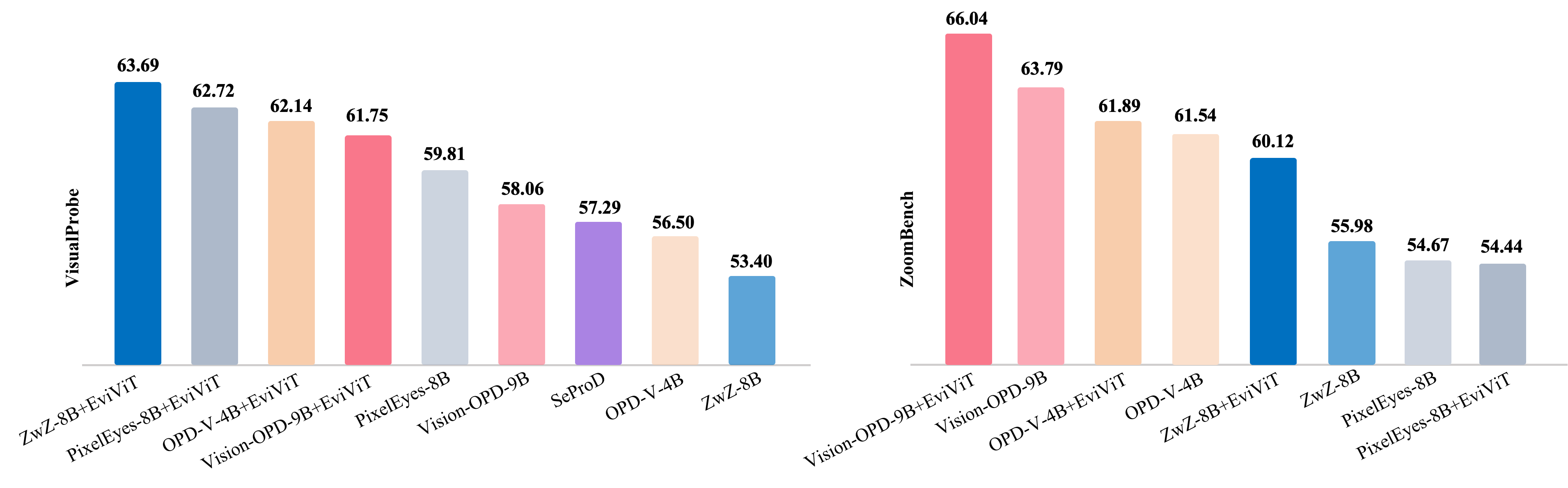}
    \caption{\textbf{EviViT advances strong models on fine-grained benchmarks.}
    ZwZ-8B with EviViT leads VisualProbe, and Vision-OPD-9B with EviViT leads
    ZoomBench among the compared open models. Both directly reuse attachments
    trained on their respective base models. Darker shades mark models with
    EviViT. We evaluate the open checkpoints locally. The SeProD bar combines
    its published scores across the three VisualProbe difficulty levels under
    avg@32.}
    \label{fig:sota_frontier}
\end{figure*}

%% file: Sec/exp.tex
\input{Tabs/tab_cross_encoder}

\section{Experiments}
\label{sec:experiments}

\paragraph{Evaluation protocol.}
The five fine-grained test sets contain 3,151 questions. VisualProbe
(VP) has 515 across three difficulty levels~\citep{lai2026mini},
V$^\ast$Bench (V$^\ast$) has 191~\citep{wu2024v}, and HR-Bench has 800 at
each resolution (4K and 8K)~\citep{wang2025divide}. We use the full image
for each of the 845 ZoomBench questions~\citep{wei2026zooming}. VP-All
pools the three VP levels. We report accuracy differences in percentage
points (pp).

Local paired runs share images, questions, language weights, prompts and
deterministic decoding. The frozen Qwen3-VL-8B judge sees only the question,
reference and candidate answer. Missing final answers count as incorrect.

\input{Tabs/tab_search_4k}
\input{Tabs/tab_budget_scaling}

\subsection{Fine-grained perception and model transfer}
\label{subsec:fine_grained}

In Table~\ref{tab:fine_grained}, EviViT improves average fine-grained
accuracy over a single full-image view (Global) on all nine models under
their native settings. VisualProbe Medium and
Hard show the largest mean gains on VP, both above 8~pp. All nine models
also improve on HR-Bench 8K and ZoomBench, showing that the search cues
learned on VisualProbe transfer to other benchmarks.

\paragraph{One attachment serves a model family.}
Five descendants reuse the four Qwen base models' attachments unchanged.
Among the open models in Figure~\ref{fig:sota_frontier}, ZwZ-8B sets the
state of the art on VisualProbe (63.69\%) and Vision-OPD-9B on
ZoomBench (66.04\%)~\citep{he2026self,yuan2026vision,bi2026opd,sarch2026vero,gong2026pixeleyes}.

The same Qwen3-VL-8B attachment raises average accuracy by about 6~pp on the base model, ZwZ-8B and Vero-Qwen3I-8B (Vero-8B).
In Supplementary Figure~\ref{fig:case_vero_reasoning}, Vero-8B calls a
sign oval from the global view alone. With EviViT, it recognizes the
sign's straight edges and right angles. ZwZ-4B and both Vision-OPD models
also gain in average accuracy by loading their base model's attachment.

\paragraph{Transfer across visual encoders.}
LLaVA-OneVision encodes fixed-size tiles with SigLIP, while
LLaVA-OneVision-2 uses the native-resolution OneVision-Encoder~\citep{li2024llava,an2026llava}.
We train a separate attachment for each, mapping the same human search
traces from image coordinates to each encoder's grid (Supplementary
Sec.~\ref{app:cross_encoder}). Under the shared ceiling of 4,096 visual
tokens (4K) per question in Table~\ref{tab:cross_encoder}, both gain more
than 6~pp in pooled accuracy over Global. Although the attachments learn
from VisualProbe searches, both models also improve on the two HR-Bench
splits and on ZoomBench. These gains show that EviViT adapts to different
visual encoders.

\paragraph{Evidence re-reading versus visual search.}
The single-call methods in Table~\ref{tab:search_4k} share the same 4K
ceiling across all views. EviViT reaches 66.90\% pooled accuracy on
Qwen3-VL-8B. On VP, the released Mini-o3 model outperforms
EviViT but submits $10.8\times$ as many visual tokens over the course
of a rollout (Supplementary Sec.~\ref{app:search_budget}). Repeated model calls and crop operations also add latency before the
final answer.
With the Qwen model held fixed, gathering evidence before generation
lets EviViT outperform both self-grounding and FOCUS in pooled accuracy
with a single answering call.

\subsection{Accuracy--efficiency trade-off}
\label{subsec:efficiency}

\paragraph{Better accuracy with fewer visual tokens.}
Both Qwen3-VL sizes improve mean VP/V$^\ast$ accuracy at all seven
ceilings in Table~\ref{tab:budget_scaling} and Figure~\ref{fig:budget_scaling},
from 1K to 20K. EviViT stays within Global's actual token count on every
question.

\begin{samepage}
On 8B, EviViT at 4K reaches 68.87\% with 3.62K visual tokens on average.
Global at 20K reaches 68.27\% with 13.31K. Selective re-reading reaches
comparable accuracy with about a quarter of the visual input.\par
\end{samepage}

\par\smallskip\noindent\begin{minipage}{\columnwidth}
\captionsetup{hypcap=false}
    \centering
    \includegraphics[width=\columnwidth]{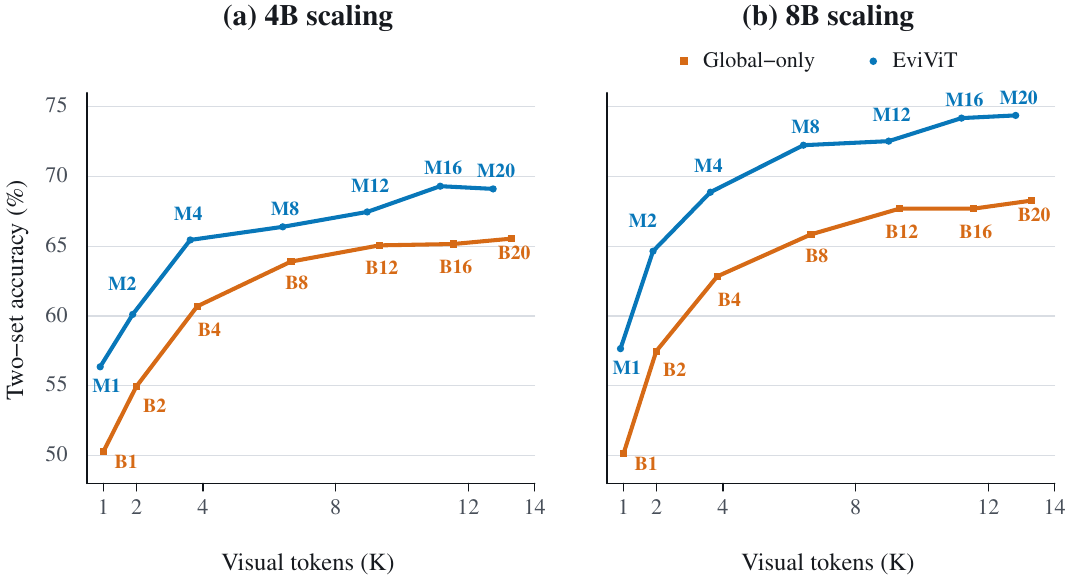}
    \captionof{figure}{\textbf{Accuracy versus realized visual tokens.} Points give mean VP/V$^\ast$ accuracy. The x-axis shows the average number of visual tokens used per question (K). Point labels give the budget ceiling in K for Global (B) and EviViT (M).}
    \label{fig:budget_scaling}
\end{minipage}\par\smallskip

\input{Tabs/tab_cost_overview}

\begin{samepage}
\paragraph{End-to-end request time.}
At the 4K and 16K ceilings in Table~\ref{tab:cost_overview}, EviViT
raises mean VP/V$^\ast$ accuracy on Qwen3-VL-8B by more than 6~pp over
Global. The median request takes less than 20\,ms longer at either ceiling (Supplementary Table~\ref{tab:efficiency}).
\par\end{samepage}

EviViT and native multi-image input use the same source crops and token
counts. We include online selection, cropping and preprocessing in both
end-to-end measurements.
The native interface encodes the global
view again to answer. EviViT reuses those features and needs no external
crop tool. Its mean request time is 5.6\% lower on 40 isolated 4K requests
(Supplementary Sec.~\ref{app:isolated_cost}). In a separate 4K accuracy test, the multi-image baseline crops the resized
global view, while EviViT re-reads the original image. With the same token
count per question, EviViT gains 1.48~pp in pooled accuracy
(Supplementary Table~\ref{tab:source_paths}).

\subsection{Preserving general capabilities}
\label{subsec:general_preservation}

EviViT learns where to look from VisualProbe searches. To see how this
training affects broader visual tasks, we evaluate the four Qwen base
models on MMBench~\citep{liu2024mmbench}, MMStar~\citep{chen2024we}
and Visual-CoT~\citep{shao2024visual} without refitting their attachments.
Ten of the twelve scores in Table~\ref{tab:compatibility_preservation}
improve. EviViT thus largely preserves the models' general visual
capabilities.

\input{Tabs/tab_compatibility_preservation}

\subsection{Continued trainability}
\label{subsec:continued_training}

Section~\ref{subsec:fine_grained} shows that a model can load EviViT after
post-training. To test whether it can continue learning with EviViT in
place, we use the same 10K examples for supervised fine-tuning (SFT)
of Qwen3-VL-8B with and without the attachment. We update only the
language-side LoRA weights. Both models improve, and the model with
EviViT retains a 5.05~pp lead in mean accuracy after SFT
(Supplementary Table~\ref{tab:sft_recovery}).

\subsection{Ablation study}
\label{subsec:main_ablation}

\paragraph{Question-aligned evidence benefits from sparse fusion.}
All four conditions in Table~\ref{tab:component_v1}\,(a) answer the
original question. With the learned Bridge, selecting regions for that
question improves V$^\ast$ accuracy by 3.14~pp over selection with a
shuffled question. On the same crops, the learned Bridge gains 1.05~pp on
V$^\ast$ over identity, which leaves the global and regional features
unchanged at this layer. Moving the crops also lowers accuracy with native multi-image
input (Supplementary Sec.~\ref{subsec:ablation}). On both VP and V$^\ast$,
selecting the right regions accounts for the larger gain, and fusion helps
the model use the evidence they contain.

\input{Tabs/tab_component_v1}

\begin{samepage}
\paragraph{What the search trace contributes.}
The two conditions in Table~\ref{tab:trace_supervision}\,(b) use the same
examples, random seed and three epochs of Bridge training. Full searches
improve all three scores over supervision from final boxes and their
expansions, with the largest gain on VP-Hard (+3.77~pp). A final box marks where the search ended. The full trace also shows
which parts of the scene the annotator inspected before narrowing in on
the answer, giving the model a richer target for learning where to look.
\par\end{samepage}

%% file: Tabs/tab_cross_encoder.tex
\begin{table}[!b]
\centering
\caption{\textbf{Transfer across visual encoders.} Accuracy (\%) with a shared ceiling of 4,096 visual tokens per question. HR4 and HR8 denote the HR-Bench resolution splits. Avg.\ pools questions across the five test sets. Bold marks the higher score within a model; $\Delta$ is the gain over Global (pp).}
\label{tab:cross_encoder}
\small
\setlength{\tabcolsep}{2.8pt}
\begin{tabular*}{\columnwidth}{@{\extracolsep{\fill}}lrrrrrr@{\hspace{4pt}}r@{}}
\toprule
Input & VP & V$^\ast$ & HR4 & HR8 & Zoom & Avg. & \multicolumn{1}{c@{}}{$\Delta$}\\
\midrule
\multicolumn{8}{@{}l}{\textit{LLaVA-OneVision-7B}}\\
Global & 20.97 & 62.30 & 53.75 & 48.13 & 37.28 & 43.07 & \\
EviViT & \textbf{34.17} & \textbf{66.49} & \textbf{59.38} & \textbf{53.13} & \textbf{41.18} & \textbf{49.22} & \gain{(+6.16)}\\
\addlinespace[3pt]
\multicolumn{8}{@{}l}{\textit{LLaVA-OneVision-2-8B}}\\
Global & 32.62 & \textbf{87.43} & 70.13 & 65.00 & 45.21 & 57.06 & \\
EviViT & \textbf{42.72} & 84.82 & \textbf{78.13} & \textbf{71.63} & \textbf{49.11} & \textbf{63.31} & \gain{(+6.25)}\\
\bottomrule
\end{tabular*}
\end{table}

%% file: Tabs/tab_search_4k.tex
\begin{table}[!b]
\centering
\caption{\textbf{Evidence re-reading and visual search.} Accuracy (\%). VP pools its three difficulty levels, and All pools the five test sets. HR4 and HR8 denote HR-Bench at 4K and 8K. Bold marks the highest score in each column. Supplementary Sec.~\ref{app:search_budget} gives each method's budget, model and search procedure.}
\label{tab:search_4k}
\small
\setlength{\tabcolsep}{2pt}
\begin{tabular*}{\linewidth}{@{\extracolsep{\fill}}lrrrrrr@{}}
\toprule
Method & VP & V$^\ast$ & HR4 & HR8 & Zoom & All\\
\midrule
\multicolumn{7}{@{}l}{\textit{One call on Qwen3-VL-8B, shared 4K ceiling}}\\
Global & 39.81 & 85.86 & 75.50 & 68.63 & 44.85 & 60.33\\
EviViT & 48.16 & \textbf{91.10} & \textbf{80.88} & \textbf{77.88} & 49.23 & \textbf{66.90}\\
\midrule
\multicolumn{7}{@{}l}{\textit{Multi-step search on Qwen3-VL-8B, 4K per call}}\\
Self-grounding & 45.44 & 81.15 & 77.63 & 73.63 & \textbf{55.50} & 65.63\\
FOCUS & 43.11 & 87.96 & 76.75 & 70.88 & 52.31 & 63.88\\
\midrule
\multicolumn{7}{@{}l}{\textit{Multi-step search on each method's released model}}\\
Mini-o3 & \textbf{52.43} & 86.91 & 74.25 & 70.00 & 48.52 & 63.47\\
SEAL & 20.97 & 74.87 & 33.00 & 37.00 & 30.30 & 33.86\\
\bottomrule
\end{tabular*}
\end{table}

%% file: Tabs/tab_budget_scaling.tex
\begin{table*}[!t]
\centering
\caption{\textbf{Accuracy--budget scaling.} Accuracy (\%) averages VP-All and V$^\ast$ equally. Each EviViT input uses at most the paired Global token count on that question. Tokens are realized means in thousands.}
\label{tab:budget_scaling}
\normalsize
\setlength{\tabcolsep}{2pt}
\renewcommand{\arraystretch}{1.07}
\begin{tabular*}{\textwidth}{@{\extracolsep{\fill}}r rrr rr @{\hspace{12pt}} rrr rr@{}}
\toprule
 & \multicolumn{3}{c}{\textbf{Qwen3-VL-4B} accuracy (\%)} & \multicolumn{2}{c}{Tokens (K)} & \multicolumn{3}{c}{\textbf{Qwen3-VL-8B} accuracy (\%)} & \multicolumn{2}{c}{Tokens (K)} \\
\cmidrule(lr){2-6}\cmidrule(l){7-11}
Ceiling & Global & EviViT & Gain (pp) & Global & EviViT & Global & EviViT & Gain (pp) & Global & EviViT \\
\midrule
1K & 50.30 & \textbf{56.35} & \gain{+6.05} & 1.00 & 0.90 & 50.14 & \textbf{57.65} & \gain{+7.51} & 1.00 & 0.91 \\
2K & 54.93 & \textbf{60.11} & \gain{+5.18} & 1.99 & 1.88 & 57.47 & \textbf{64.64} & \gain{+7.17} & 1.99 & 1.89 \\
4K & 60.69 & \textbf{65.45} & \gain{+4.76} & 3.83 & 3.62 & 62.83 & \textbf{68.87} & \gain{+6.04} & 3.83 & 3.62 \\
8K & 63.89 & \textbf{66.39} & \gain{+2.49} & 6.65 & 6.41 & 65.84 & \textbf{72.24} & \gain{+6.40} & 6.65 & 6.43 \\
12K & 65.06 & \textbf{67.46} & \gain{+2.40} & 9.32 & 8.95 & 67.69 & \textbf{72.53} & \gain{+4.84} & 9.32 & 9.00 \\
16K & 65.16 & \textbf{69.30} & \gain{+4.15} & 11.55 & 11.15 & 67.69 & \textbf{74.18} & \gain{+6.49} & 11.55 & 11.19 \\
20K & 65.54 & \textbf{69.11} & \gain{+3.56} & 13.31 & 12.75 & 68.27 & \textbf{74.38} & \gain{+6.10} & 13.31 & 12.83 \\
\bottomrule
\end{tabular*}
\end{table*}

%% file: Tabs/tab_cost_overview.tex
\par\smallskip\noindent\begin{minipage}{\columnwidth}
\captionsetup{hypcap=false}
\centering
\captionof{table}{\textbf{Latency and accuracy at three token ceilings.} Median request time (s) after warm-up for Global $\rightarrow$ EviViT. Gain is the mean change in VP/V$^\ast$ accuracy (pp).}
\label{tab:cost_overview}
\small
\renewcommand{\arraystretch}{1.12}
\setlength{\tabcolsep}{1.8pt}
\begin{tabular*}{\columnwidth}{@{\extracolsep{\fill}}r c r @{\hspace{9pt}} c r@{}}
\toprule
 & \multicolumn{2}{c}{\textbf{Qwen3-VL-4B}} & \multicolumn{2}{c}{\textbf{Qwen3-VL-8B}} \\
\cmidrule(lr){2-3}\cmidrule(l){4-5}
Ceiling & Time (s) & Gain (pp) & Time (s) & Gain (pp) \\
\midrule
1K & 0.557\,\(\rightarrow\)\,0.738 & \gain{+6.05} & 0.696\,\(\rightarrow\)\,0.892 & \gain{+7.51} \\
4K & 0.921\,\(\rightarrow\)\,0.999 & \gain{+4.76} & 1.087\,\(\rightarrow\)\,1.105 & \gain{+6.04} \\
16K & 0.902\,\(\rightarrow\)\,1.082 & \gain{+4.15} & 1.088\,\(\rightarrow\)\,1.091 & \gain{+6.49} \\
\bottomrule
\end{tabular*}
\end{minipage}\par\smallskip

%% file: Tabs/tab_compatibility_preservation.tex
\begin{table}[!htb]
\centering
\caption{\textbf{General capabilities on four Qwen base models.} Base $\to$ EviViT accuracy (\%); the EviViT score is green for a gain and red for a loss.}
\label{tab:compatibility_preservation}
\small
\setlength{\tabcolsep}{0pt}
\renewcommand{\arraystretch}{1.12}
\begin{tabular}{@{}>{\raggedright\arraybackslash}p{.247\linewidth}>{\raggedleft\arraybackslash}p{.251\linewidth}>{\raggedleft\arraybackslash}p{.251\linewidth}>{\raggedleft\arraybackslash}p{.251\linewidth}@{}}
\toprule
Model & MMBench & MMStar & Visual-CoT \\
\midrule
Qwen3-VL-4B & 87.36\,$\to$\,\gain{88.43} & 61.73\,$\to$\,\gain{62.80} & 77.53\,$\to$\,\loss{75.52} \\
Qwen3-VL-8B & 88.59\,$\to$\,\gain{88.94} & 65.07\,$\to$\,\gain{65.27} & 77.92\,$\to$\,\gain{78.35} \\
Qwen3.5-4B & 86.37\,$\to$\,\gain{87.76} & 63.20\,$\to$\,\gain{64.40} & 77.18\,$\to$\,\gain{78.17} \\
Qwen3.5-9B & 87.13\,$\to$\,\loss{86.76} & 64.33\,$\to$\,\gain{65.20} & 78.98\,$\to$\,\gain{80.34} \\
\bottomrule
\end{tabular}
\end{table}

%% file: Tabs/tab_component_v1.tex
\begin{table}[!htb]
\centering
\caption{\textbf{Ablation study on Qwen3-VL-8B.} Accuracy (\%). In (a), parentheses give changes in pp from aligned routing with the learned Bridge. Panel (b) compares independent fits after three epochs of Bridge training. Differences are computed before rounding.}
\label{tab:component_v1}\label{tab:trace_supervision}
\small
\renewcommand{\arraystretch}{1.12}
\begin{tabular*}{.96\columnwidth}{@{\extracolsep{\fill}}l l r@{\,}l r@{\,}l@{}}
\toprule
\multicolumn{6}{@{}l}{\textbf{(a) Question routing and fusion}}\\
Routing & Bridge & \multicolumn{2}{c}{VP-All} & \multicolumn{2}{c}{V$^\ast$}\\
\midrule
\smash{\rlap{\color{basegrey}\rule[-3pt]{.96\columnwidth}{11.7pt}}}Aligned & Learned & \textbf{57.48} &  & \textbf{91.10} & \\
Aligned & Identity & 56.89 & \loss{($-0.58$)} & 90.05 & \loss{($-1.05$)}\\
Shuffled & Learned & 53.01 & \loss{($-4.47$)} & 87.96 & \loss{($-3.14$)}\\
Shuffled & Identity & 53.40 & \loss{($-4.08$)} & 86.91 & \loss{($-4.19$)}\\
\midrule
\end{tabular*}\par
\input{Tabs/tab_trace_supervision}
\end{table}

%% file: Tabs/tab_trace_supervision.tex
\begin{tabular*}{.96\columnwidth}{@{\extracolsep{\fill}}l r r r@{}}
\multicolumn{4}{@{}l}{\textbf{(b) Spatial supervision}}\\
Supervision & VP-All & VP-Hard & V$^\ast$\\
\midrule
Final box & 56.12 & 52.83 & 89.01\\
\smash{\rlap{\color{basegrey}\rule[-3pt]{.96\columnwidth}{11.7pt}}}Full trace & \textbf{57.28} & \textbf{56.60} & \textbf{90.05}\\
Difference (pp) & \gain{+1.17} & \gain{+3.77} & \gain{+1.05}\\
\bottomrule
\end{tabular*}

%% file: Sec/conclusion.tex
\section{Conclusion and limitations}
\label{sec:conclusion}

\method{} learns from human searches where to re-read the original image.
At test time, the encoder needs only the image and question to gather
evidence from several regions before the language model answers in a
single call.
Joining regional detail with the global scene improves average fine-grained
accuracy on all nine Qwen-family models. Compatible descendants reuse their
base model's attachment. With the attachment frozen, language fine-tuning
improves mean accuracy and preserves EviViT's lead over the fine-tuned
baseline. The OneVision results show
that the same search traces can guide training on other visual encoders.

At every tested budget, EviViT outperforms the global baseline and
uses fewer visual tokens on average. Reusing global features limits the
added latency. Accuracy on general visual tasks remains broadly stable.

\paragraph{Limitation \& future work.}
Collecting searches across a wider range of objects and scenes could
extend EviViT to small-object and embodied perception, where the model
must interpret local details in the wider scene.

%% file: Sec/statements.tex
\section*{LLM Usage Statement}
We used OpenAI GPT and Codex to help polish the manuscript, plan
experimental comparisons, and write code for experiments, analysis and
figures. A frozen Qwen3-VL-8B judged answers. Table~\ref{tab:fine_grained}
marks the GPT and Gemini scores we obtained via API. The authors reviewed
the text, code and results and take responsibility for the work.

%% file: Sec/appendix.tex
\section{Reproducibility details}
\label{app:reproducibility}

\subsection{Model and training instantiation}
\label{app:frozen_models}
\paragraph{Frozen model.}
The Qwen3-VL-4B vision transformer (ViT) has 24 blocks of width 1,024
and receives \method{} after
Block 16. Qwen3-VL-8B has 27 blocks of width 1,152 and receives it after Block
18. Both scales use the same architecture, with separate weights for the
Prompt-conditioned Token--Evidence Aligner (PTEA), Evidence Region Planner
and Sparse Bridge. The ViT, merger and large language model (LLM) remain frozen during all
four stages.
For Qwen3.5-4B and Qwen3.5-9B, we fit separate attachments on the frozen models
after visual Blocks 16 and 18, respectively; each attachment is reused on the
corresponding Vision-OPD descendant without refitting.

\subsection{Attachment architecture}
\label{app:attachment_settings}

\paragraph{Search targets and loss.}
In Eq.~\eqref{eq:trace_map}, the weights $w_k$ are 0.25 for the whole
search, 0.35 for the branch after the last reset, 0.25 for the last
completed zoom and 0.15 for the final box. In Eq.~\eqref{eq:map_loss},
we set $\lambda_1=0.2$ for the cosine loss and $\lambda_2=0.1$ for
probability mass on abandoned regions.

\paragraph{PTEA.}
Frozen language embeddings are reduced to 512 dimensions by a fixed random
projection and then to $d=192$ by a learned projection. A one-layer, four-head
text Transformer (48 dimensions per head, $192\!\rightarrow\!384\!\rightarrow
\!192$ feed-forward network, pre-normalization, and dropout) contextualizes the
question. Each visual position concatenates its projected feature, aligned text
context, elementwise product and absolute difference, maximum similarity,
attention entropy, and a learned coordinate embedding. Two $3\!\times\!3$
convolutions and a $1\!\times\!1$ head produce the evidence logits. The 4B
evidence branch contains approximately 1.41M parameters. A complementary branch
shares the visual--text alignment and has its own spatial residual block and
prediction head for $P_{\mathrm{ctx}}$. Its normalized target combines a map
of dwell and zoom events across the full trace with a map of inspections before
the last reset in a 0.70/0.30 mixture; when the latter is empty, it uses the
full-trace exploration map alone. The context
branch is initialized from PTEA and fitted with the shared alignment and
evidence branch frozen.

\paragraph{Evidence Region Planner.}
Residual suppression proposes two decisive anchors but discards the second when
it is redundant with the first. The context map is masked at every retained
decisive anchor and supplies one context region. The deployed model therefore
uses two or three regions under a three-slot cap. Box centers and extents are
normalized by the source-image width and height. For anchor $b=(c_x,c_y,w,h)$,
the geometry head predicts $(\delta_x,\delta_y,\delta_w,\delta_h)$ with a
residual multilayer perceptron (MLP; layer normalization, then widths 256, 128 and 4).
Write $(c,s)$ for the center and extent on one axis. On the horizontal axis
$(c,s,\delta_c,\delta_s)=(c_x,w,\delta_x,\delta_w)$, and the vertical axis
uses $(c_y,h,\delta_y,\delta_h)$. On each axis the updated center is $c'=c+1.5s\tanh(\delta_c)$.
The extent changes multiplicatively and is clipped to the allowed range:
\begin{equation}
  s'=\operatorname{clip}_{[0.02,\,0.95]}
      \!\left(s\,e^{1.2\tanh(\delta_s)}\right).
\end{equation}
The zero-initialized output makes the initial transform an identity. Smooth-$L_1$,
generalized intersection over union (GIoU)~\citep{rezatofighi2019generalized}, target-coverage, excess-area, and
identity losses fit the bounded
correction. The budget head is a 129-parameter MLP with six inputs and a 16-unit
hidden layer. Its inputs are the secondary decisive-region mass, the normalized entropy
of each map, their Jensen--Shannon divergence, context mass outside the decisive
regions, and peak evidence probability. It predicts a target context share
between 0.10 and 0.25, leaving the remainder for decisive views. These shares
set local allocation priorities; the final token counts also depend on region
size, the available capacity, and patch-grid alignment.

\paragraph{Sparse Bridge.}
The 4B Bridge has one block, internal width 256, four attention heads, and
approximately 1.91M parameters. Each local token attends to a $3\times3$
neighborhood centered on its global parent. Relative offset and log scale pass through a
two-layer projection to form the attention bias. Output projections are
zero-initialized, and each residual update is clipped to 20\% of the receiving
token's root-mean-square magnitude. Reverse messages are aggregated by coordinate parent and written
only to global tokens that have local children.

\paragraph{Four-stage fit.}
First, we cache frozen ViT features at candidate insertion blocks for the
annotated VisualProbe (VP) training examples. Five-fold cross-validation using only the trace
targets selects the insertion block and PTEA training duration before external
question answering (QA) evaluation. Second,
PTEA is optimized for three complete passes over the cached features using
Equation~\ref{eq:map_loss}, after which the complementary context branch is
fitted with PTEA frozen. Third, the planner's budget and geometry heads are fitted from the
trace-derived context-share target and last-zoom/final-box geometry. The geometry head initially leaves the proposed boxes unchanged because its
output layer starts at zero.
Fourth, with routing fixed, the Sparse Bridge is trained with reference-answer
token cross-entropy and a 0.01 identity penalty. We evaluate the checkpoint
saved after the first complete pass. At inference, a fixed rule converts the planner's predicted shares and view sizes into token grids.

Table~\ref{tab:bridge_configuration} summarizes the experimental setup and
the Bridge settings for the four Qwen base attachments. Independent runs
share the server; each Bridge fit uses one GPU. The global and local token
budgets in the table apply to Bridge fitting. On Qwen3-VL-8B, residual peaks of the evidence map select up to three training crops that share the local budget. At evaluation, the same weights process the grids assigned by the planner under each experiment's budget.
\input{Tabs/tab_bridge_configuration}

\begin{figure*}[t]
    \centering
    \includegraphics[width=0.95\linewidth]{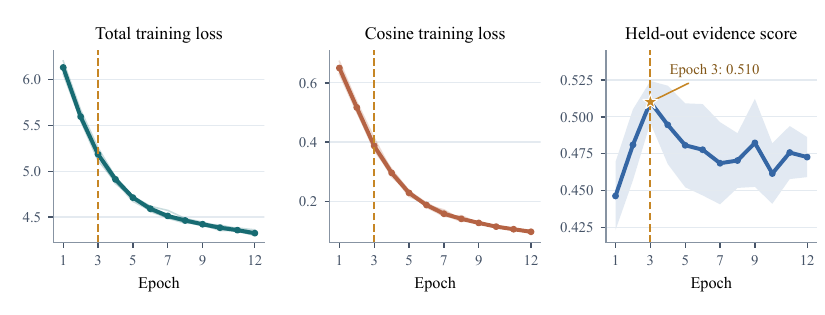}
    \caption{\textbf{PTEA training and checkpoint selection.} Five-fold
    trace-only training of the Qwen3-VL-4B Block-16 selector. Thin curves show
    individual folds and thick curves their mean. The held-out evidence score
    is shown with a one-standard-deviation band across folds; dashed lines mark
    the selected third epoch.}
    \label{fig:ptea_training_cv}
\end{figure*}

Training loss alone does not identify PTEA's best spatial cue. In the 4B
Block-16 five-fold run, both losses keep falling through epoch twelve, while
the held-out evidence score in Figure~\ref{fig:ptea_training_cv} peaks at epoch
three. The score reflects map alignment and coverage of final and last-zoom
regions. We therefore fit a separate PTEA on the full set of traces for three
epochs. Only the held-out trace targets guide this choice.

For the 4B core, frozen feature extraction, three PTEA passes and the
selected Bridge pass total 1.046 A100 GPU-hours in the training logs. Peak
process memory was 36,993 MiB. These stage timings were collected under
shared-machine loads and exclude geometry-head calibration.

\paragraph{Token allocation under native settings.}
\label{app:capacity_floor}
Let $N$ be the post-merger token count under the model's original 16M-pixel
processor. The planner uses this count to set the scale
\begin{equation}
  F(N)=\operatorname{clip}_{[2048,\,4096]}(1024+N/4).
  \label{eq:native_floor}
\end{equation}
It combines this scale with native demand through the capacity envelope
$\sqrt{N^2+F(N)^2}$. The global view is planned from image size and native
capacity; regional views receive the remaining capacity according to their
area and allocation priorities, including the planner's predicted context share.
Each allocation is then aligned to the model patch grid, so $F(N)$ is a soft
capacity parameter rather than a guaranteed minimum realized token count.
The same rule applies across datasets. In the matched-budget sweep, an explicit
ceiling instead caps each sample at the realized token count of its paired
global-only baseline. This ceiling overrides the soft-capacity envelope when
necessary, including every sample at 1K and larger images at 2K. All reported
visual-token counts are measured after grid alignment.

\subsection{Isolated latency and memory}
\label{app:isolated_cost}

Table~\ref{tab:efficiency} reports the time to first token (TTFT), end-to-end (E2E) latency,
and peak memory for paired Global and EviViT settings. Each setting uses
64 isolated requests. The first four warm up the model, and the remaining
60 determine the median and 90th percentile (P90). We measure accuracy
gains in percentage points (pp) on the full VP and V$^\ast$ sets, as in
Table~\ref{tab:budget_scaling}.
\input{Tabs/tab_efficiency}

In the separate native multi-image timing comparison
(Sec.~\ref{subsec:efficiency}), we measure 40 requests on the same isolated
GPU, from opening the original image to finishing the answer. Each path
uses a resident model and one separate warm-up. Both select the same crops
online from the original pixels and submit equal visual-token counts on
every question. The native path reuses preprocessed pixels; EviViT also
reuses the global features already computed for selection. The
thumbnail-crop control in Supplementary Table~\ref{tab:source_paths} is
an accuracy comparison, not the baseline for this timing measurement.

\subsection{Visual-search protocols}
\label{app:search_budget}
Table~\ref{tab:search_4k} uses the same questions, reference answers and
Qwen3-VL-8B judge across methods. A per-call ceiling counts the visual
positions from \emph{all} views submitted in that call. Summing those counts
over the question also counts history resubmitted on later calls.

\paragraph{One call on Qwen3-VL-8B.}
Global allocates up to 4,096 visual tokens to a single image. EviViT shares
the same ceiling across the global view and all re-read regions.

\paragraph{Repeated calls on Qwen3-VL-8B.}
Self-grounding uses a global view to predict boxes, then shares a fresh 4K
allowance across the global view and the crops for answering. The two calls
average 7,833 submitted visual tokens per question. FOCUS~\citep{zhong2026focus}
uses the model to select and inspect high-scoring regions before answering.
Its search and answer calls each use the shared 4K ceiling.

\paragraph{Released search systems.}
Mini-o3~\citep{lai2026mini} uses its released answering weights, crop tool and
greedy rollout. The initial image and each new crop independently receive up
to 4,096 visual tokens. The model keeps earlier views in a context of 32,768 tokens for up to
12 rounds. Across the five test sets, it averages 3.69 language model
calls and 25,698 submitted visual tokens per question, including history.
On VP, Mini-o3 averages 5.32 calls and 43,596 submitted visual tokens per
question including history, versus one call and 4,025 tokens for EviViT.
The V$^\ast$ search system, SEAL~\citep{wu2024v}, uses its released 7B search and visual question answering models with native
image resolutions and object-feature selection. Each language prefill,
including the final answer context, is limited to 4,096 inserted visual
positions. Its separate vision-encoder passes also contribute computation.
Input counts throughout this comparison measure submitted visual positions.
Elapsed time requires a separate measurement.

\subsection{Adaptation to other visual encoders}
\label{app:cross_encoder}
For both attachments in Table~\ref{tab:cross_encoder}, the global view
and all regional views together use at most 4,096 visual tokens per question. Grid rounding and each model's legal layouts determine the realized
number of tokens.

\paragraph{LLaVA-OneVision-7B.}
The frozen SigLIP tower uses $384\times384$ views, learned absolute positions
and $14\times14$ patches. Its $27\times27$ grid covers 378 pixels per side.
We train PTEA on the block-18 global grid, mapping patch centers to source-image
coordinates. The planner assigns the available local tiles to the selected
regions. Each source-image crop is processed as a native $384\times384$ view. After the Bridge,
the model runs the remaining visual layers and its projector; image spans, newline
tokens and one-dimensional language positions use native multi-image packing.
One global plus four local tiles cost 737,280 encoded pixels and 3,650 submitted
tokens. The capped global AnyRes control averages 732,507 pixels and 2,906
tokens because it must choose a legal tile layout.

\paragraph{LLaVA-OneVision-2-8B.}
The native-resolution OneVision-Encoder has 24 visual blocks and groups raw
patches in $2\times2$ merge order. At block 16 we pool neighboring patches
only to construct PTEA's map grid; the Bridge reads and writes the unmerged
patch states using source coordinates. Each view retains its native rotary
positions and sequence boundaries through blocks 17--24 and the frozen
merger/projector. Across the five test sets, Global and EviViT average
4,061 and 4,044 submitted visual tokens per question, respectively.

For each encoder, we fit its own PTEA, planner and Bridge on the same
set of search traces. Before training the Bridge, we verify that zero
updates reproduce the native output and that gradients from the answer
loss reach the Bridge. Each Bridge fit takes 1,144 microsteps (286 updates).
During training, residual peaks in the evidence map determine the proposed
source-image crops.

\subsection{Component comparisons}
\label{subsec:ablation}

Table~\ref{tab:component_v1}\,(a) uses the Qwen3-VL-8B attachment and the
native allocation of Table~\ref{tab:fine_grained}. In the shuffled condition,
the same donor question enters PTEA and the planner's question features;
the answering model always receives the original question. Within each
routing condition, the learned and identity arms retain the same regions,
grids and input order. Identity disables the Bridge updates and preserves
both the global and regional streams. With aligned routing the Bridge gains
three correct answers on VP and two on V$^\ast$.

In Table~\ref{tab:position_controls}, randomly translating native multi-image
crops or swapping their centers across images lowers pooled accuracy by about
9.5~pp at fixed view counts, sizes and budgets.
With the same selected crops and visual token counts, EviViT matches the
accuracy of native multi-image input on VP and V$^\ast$ combined.
In the separate 4K timing test, reusing global features reduces mean
end-to-end time by 5.6\% relative to native multi-image input
(Sec.~\ref{app:isolated_cost}).

\input{Tabs/tab_position_controls}

\paragraph{Evidence localization and visual features.}
We annotated V$^\ast$ separately from the VisualProbe training data.
Five of its 191 questions had ambiguous targets, leaving 186 for localization
and coverage. All 191 count toward QA accuracy. On this subset, PTEA
distinguishes annotated pixels from background with an area under the
receiver operating characteristic curve (AUROC) of 0.9294 and average
precision of 0.2272. Its top eight proposals cover at least 90\% of every
target on 70.43\% of questions. Supplementary Sec.~\ref{app:deployed_coverage}
evaluates the two or three crops actually used for answering.

To see how much the visual features change, we also measure centered kernel
alignment between the original and augmented paths. Their representations
remain similar at insertion (0.948), at the final ViT block (0.895) and
after the merger (0.926).

Figure~\ref{fig:regional_features} compares features from visual block~19.
The clock face covers about $69\times43$ patches in the global view and
$138\times86$ when re-read from the original image.

\input{Tabs/tab_paired_audit}

Table~\ref{tab:paired_audit} tracks which answers change after adding
EviViT. At every VisualProbe difficulty level, each of the four base
models corrects more answers than it loses. On Medium, accuracy rises by
8.96--11.19~pp while the models retain 86.73--92.44\% of their previously
correct answers. The paired differences on this tier give exact McNemar
$p\leq7.2\!\times\!10^{-4}$ for all four models.

\subsection{Prompts, decoding, and scoring}
Global and EviViT start from the same source image and question in each
pair. Both use the same direct-answer prompt and deterministic decoding.
Neither receives annotated evidence boxes.

The frozen Qwen3-VL-8B judge sees only the question, reference and
candidate answer, so it can credit equivalent wording. All questions
count toward accuracy, including missing final answers, which score zero.
String matching serves as a diagnostic. For the longer responses from
OPD-V's released policy, we use a 64K-token judge context and deterministic
Flash attention for both runs. This preserves the response while keeping the
judge and its instructions fixed.

\subsection{Supervised fine-tuning (SFT)}
\label{app:recovery_sft}

\input{Tabs/tab_sft_recovery}

Table~\ref{tab:sft_recovery} uses a shared 10,000-example QA manifest with
6,694 document and optical character recognition (OCR) examples, 2,900 examples from the five other Visual-CoT
tasks, and 406 TextCaps/V7W replay examples. The document/OCR counts are
DocVQA 754, InfographicsVQA 2,004, TextVQA 2,003, DUDE 1,681, and SROIE 252.
Both arms consume only the image, question, and answer; evidence boxes are
not additional policy inputs.

Low-rank adaptation (LoRA) of the language attention projections $q,k,v,o$ uses rank 8, $\alpha=16$, and dropout
0.05. AdamW uses a learning rate of $10^{-6}$, $\beta=(0.9,0.95)$,
$\epsilon=10^{-8}$, zero weight decay, 3\% warm-up, and cosine decay.
The micro-batch is one question with 20-step gradient accumulation: each epoch
contains 10,000 sample steps and 500 optimizer updates. Both arms train for
three epochs in the same seeded order, saving every half epoch; the reported
pair uses the common final endpoint, not a separate best checkpoint per task.
The vision tower and all EviViT modules remain frozen, while language LoRA
learns from the visual tokens produced by the corresponding path.

\begin{figure*}[t]
    \centering
    \includegraphics[width=0.95\linewidth]{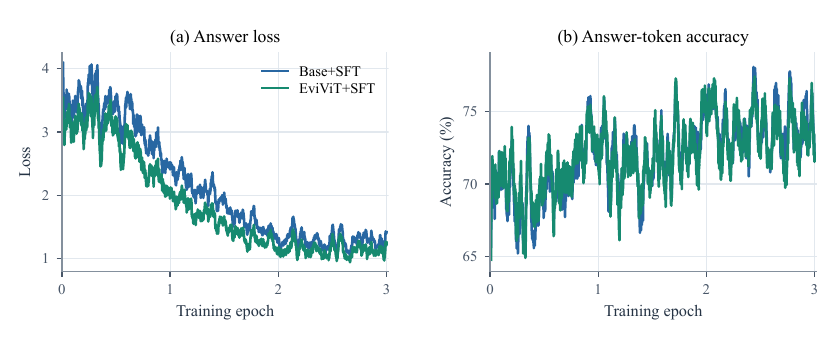}
    \caption{\textbf{Language fine-tuning with and without EviViT.} The two Qwen3-VL-8B
    runs see the same examples in the same order. The curves show answer
    loss (left) and answer-token accuracy (right), averaged over a moving
    window of 400 examples. Only language LoRA is updated.}
    \label{fig:sft_training_curves}
\end{figure*}

Visual-CoT pools 7,427 questions. Doc/OCR covers DocVQA, InfographicsVQA,
TextVQA, DUDE and SROIE. The other five tasks are Flickr30K, GQA,
OpenImages, VSR and CUB. Each group weights its tasks equally.
VP-All pools the Easy, Medium and Hard questions (141, 268 and 106,
respectively). The SFT audit uses separately
evaluated frozen controls, with the same answer prompt and judge before and
after training and judge input/output limits of 8,192/16 tokens. On the same questions, the
compatibility audit uses the earlier limits of 1,024 input and 8 output tokens.

In Table~\ref{tab:sft_recovery}, language-side SFT retains EviViT's
advantages on VisualProbe and ZoomBench while improving Visual-CoT. The
attachment stays frozen throughout the training in Figure~\ref{fig:sft_training_curves},
so the model adapts to new supervision without refitting its visual evidence path.

\input{Sec/research_statements}

\section{Additional evaluation results}
\label{app:evaluation}

\paragraph{Visual-CoT task breakdown.}
Table~\ref{tab:visual_cot_taskwise} breaks down Visual-CoT for the two Qwen3.5
base models. Both improve on Flickr30K, OpenImages, VSR and CUB. The final
row pools questions across all ten tasks.

\input{Tabs/tab_visual_cot_taskwise}

\input{Sec/appendix_evidence_audits}


\section{Qualitative evidence-localization cases}
\label{app:qualitative_cases}

\begin{figure*}[p]
    
    \centering
    \includegraphics[width=0.94\linewidth]{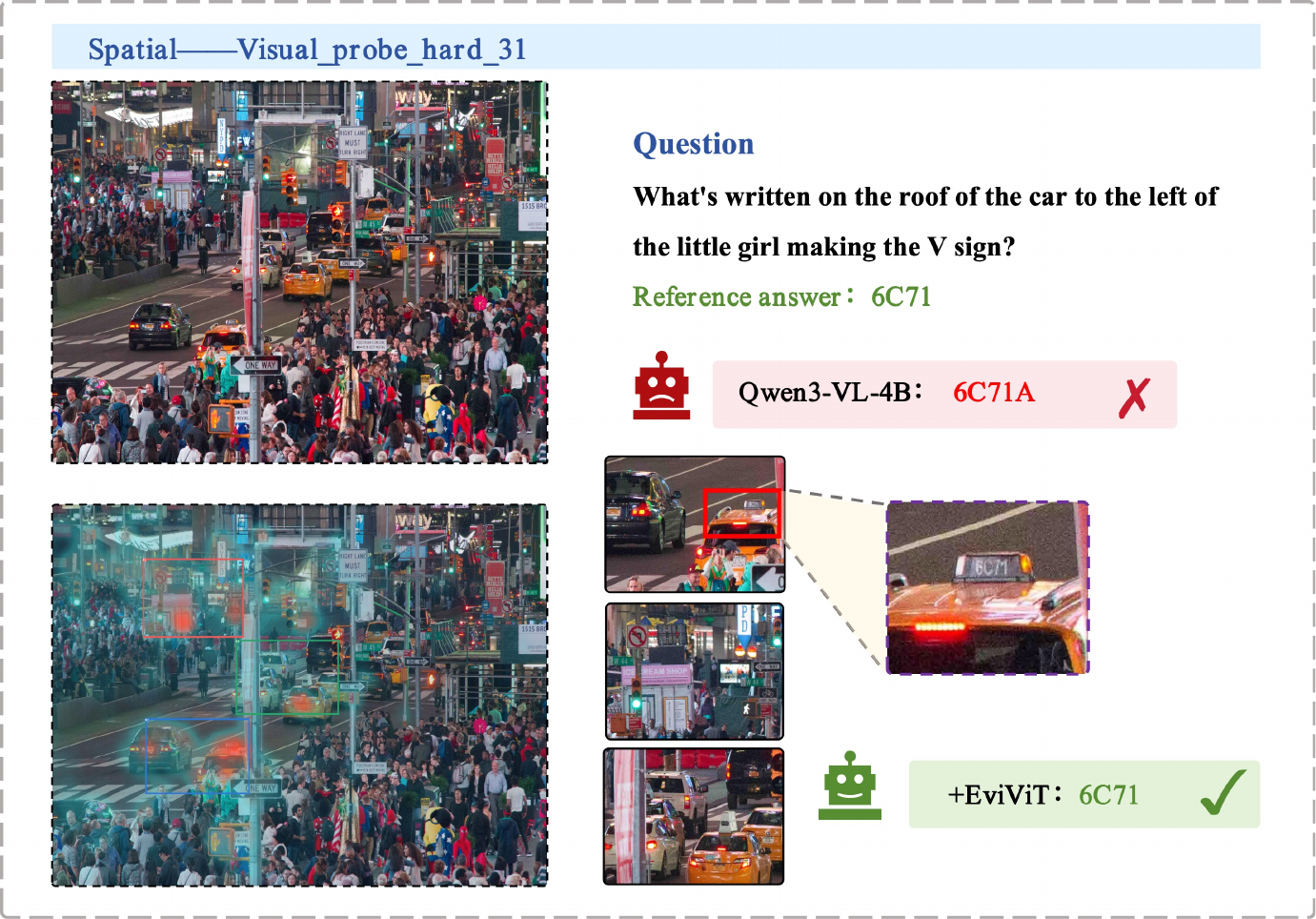}
    \caption{\textbf{Small-text repair on VisualProbe Hard.}
    The base model misreads the roof identifier as ``6C71A.'' EviViT recovers
    ``6C71'' while retaining the street context needed to locate the queried car.}
    \label{fig:case_visualprobe_ocr}
\end{figure*}

\begin{figure*}[p]
    
    \centering
    \includegraphics[width=0.94\linewidth]{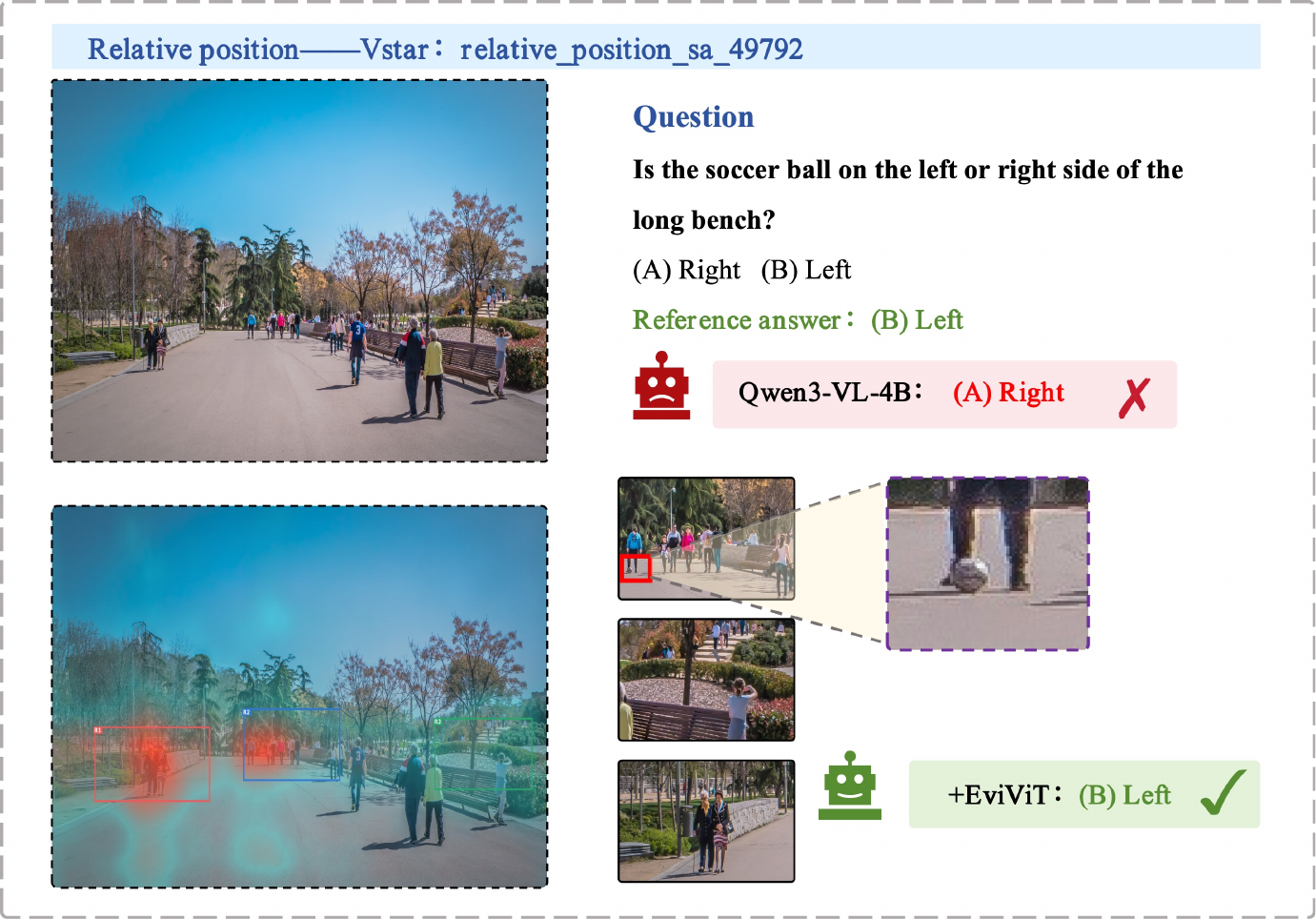}
    \caption{\textbf{Relative-position repair on V$^{\ast}$Bench.}
    Qwen3-VL-4B predicts that the soccer ball lies to the right of the long
    bench. EviViT allocates native detail to the ball, the relevant bench, and a
    complementary scene anchor, changing the answer to the correct left-side
    relation.}
    \label{fig:case_vstar_relative_position}
\end{figure*}

\begin{figure*}[p]
    
    \centering
    \includegraphics[width=0.94\linewidth]{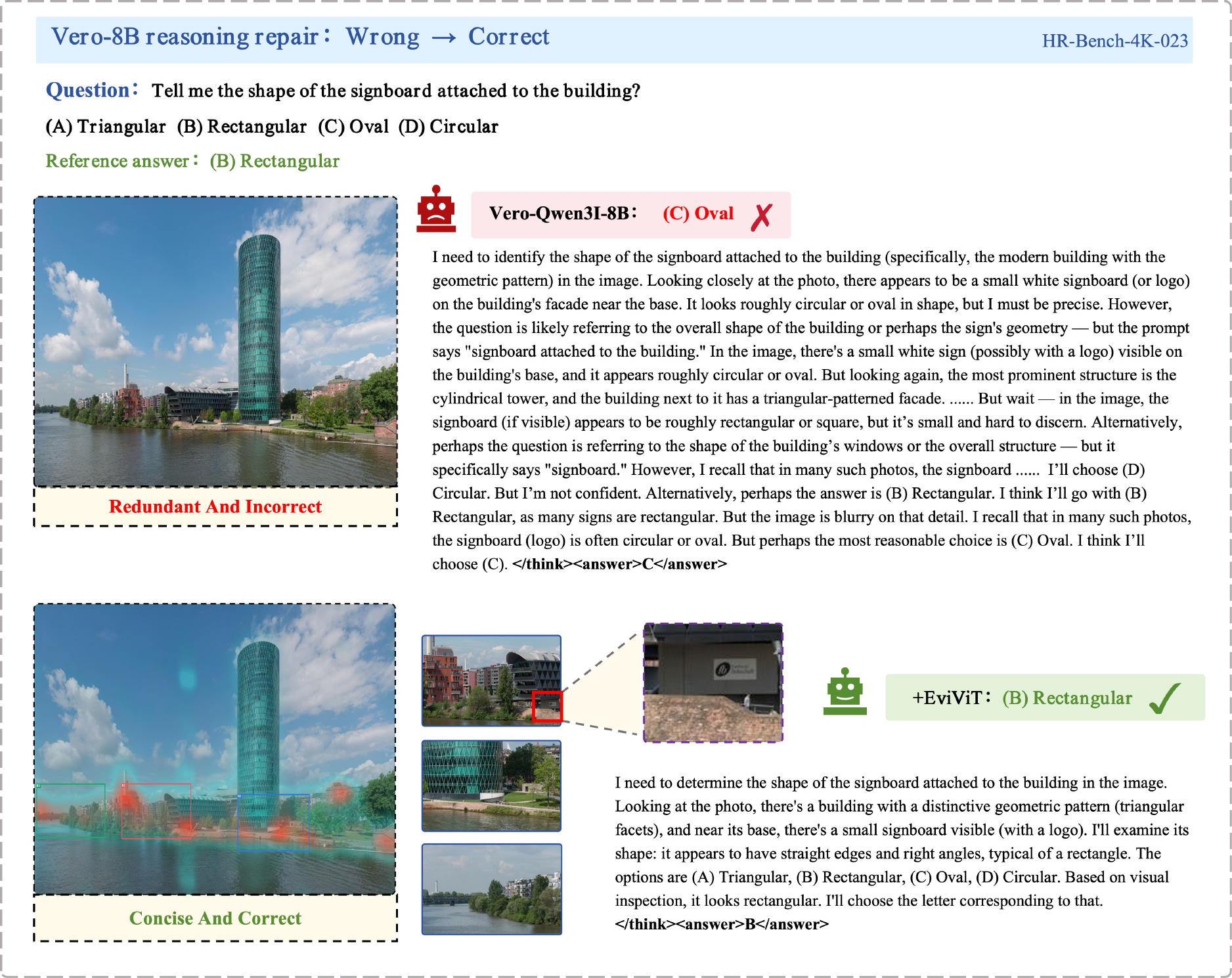}
    \caption{\textbf{Evidence-grounded reasoning repair on Vero-8B.}
    The base response considers the tower, facade, and logo before incorrectly
    answering ``oval.'' EviViT localizes the signboard and grounds a shorter,
    correct rationale in its straight edges and right angles. Both outputs
    are reproduced in full.}
    \label{fig:case_vero_reasoning}
\end{figure*}

Figures~\ref{fig:case_visualprobe_ocr}--\ref{fig:case_vero_reasoning} make the
mechanism concrete across small-text recognition, relative-position reasoning,
and evidence-grounded explanation. In each case, EviViT recovers the decisive
local cue while preserving the scene needed to bind that cue to the question.
The Vero-8B example includes both complete generated rationales: the model
speculates about several shapes, while the attachment-assisted answer refers
to the signboard\textquoteright{}s straight edges and right angles.

\par\medskip

\par\medskip

\par\medskip
\par

%% file: Tabs/tab_bridge_configuration.tex
\begin{table}[t]
\centering
\caption{\textbf{Experimental setup and compute.} Bridge settings apply to the four Qwen base attachments. LR denotes learning rate; CE denotes cross-entropy.}
\label{tab:bridge_configuration}
\small
\begin{tabular*}{\columnwidth}{@{\extracolsep{\fill}}ll@{}}
\toprule
Setting & Value \\
\midrule
Search supervision & 1,144 VP-train traces\\
Frozen model modules & ViT, merger and LLM\\
Local answer judge & Qwen3-VL-8B\\
Experiment server & $8\times$ A100-SXM4 (80\,GB each)\\
GPUs per Bridge fit & 1\\
\midrule
\multicolumn{2}{@{}l}{\textit{Qwen Bridge training}}\\
Trainable weights & Sparse Bridge \\
Global / local token budgets & 1,024 / 3,072 \\
Optimizer / weight decay & AdamW / 0.01 \\
LR, Qwen3-VL / 3.5 & $10^{-4}$ / $2\times10^{-5}$ \\
Loss & Answer CE + 0.01 residual \\
Microbatch / accumulation & 1 / 4 \\
Selected update & 286 (one pass) \\
\bottomrule
\end{tabular*}
\end{table}

%% file: Tabs/tab_efficiency.tex
\begin{table*}[!t]
\centering
\caption{\textbf{Isolated end-to-end efficiency.} Each budget pairs Global and EviViT. Token counts and accuracy gains (pp) use the full sweep. Latency statistics exclude warm-up requests.}
\label{tab:efficiency}
\tablebodyfont
\setlength{\tabcolsep}{3.4pt}
\renewcommand{\arraystretch}{1.03}
\begin{tabular}{clrrrrrrr}
\toprule
& & & \multicolumn{2}{c}{TTFT (s)} & \multicolumn{2}{c}{E2E (s)} & & \\
\cmidrule(lr){4-5}\cmidrule(lr){6-7}
Budget & Path & Vis. tok. (K) & Median & P90 & Median & P90 & Peak (GiB) & $\Delta$ Acc. (pp) \\
\midrule
\multicolumn{9}{l}{\textit{Qwen3-VL-4B}} \\
1K & Global & 1.00 & 0.206 & 0.218 & 0.557 & 0.619 & 8.62 & -- \\
\rowcolor{basegrey} & EviViT & 0.90 & 0.384 & 0.403 & 0.738 & 0.804 & 8.66 & \gain{+6.05} \\
4K & Global & 3.83 & 0.559 & 0.643 & 0.921 & 1.038 & 9.48 & -- \\
\rowcolor{basegrey} & EviViT & 3.62 & 0.621 & 0.712 & 0.999 & 1.091 & 9.47 & \gain{+4.76} \\
16K & Global & 11.55 & 0.541 & 0.637 & 0.902 & 1.021 & 9.97 & -- \\
\rowcolor{basegrey} & EviViT & 11.15 & 0.643 & 0.739 & 1.082 & 1.192 & 9.96 & \gain{+4.15} \\
\midrule
\multicolumn{9}{l}{\textit{Qwen3-VL-8B}} \\
1K & Global & 1.00 & 0.254 & 0.264 & 0.696 & 0.706 & 16.72 & -- \\
\rowcolor{basegrey} & EviViT & 0.91 & 0.465 & 0.477 & 0.892 & 0.904 & 16.76 & \gain{+7.51} \\
4K & Global & 3.83 & 0.742 & 0.887 & 1.087 & 1.231 & 17.66 & -- \\
\rowcolor{basegrey} & EviViT & 3.62 & 0.759 & 0.853 & 1.105 & 1.200 & 17.71 & \gain{+6.04} \\
16K & Global & 11.55 & 0.740 & 0.887 & 1.088 & 1.233 & 18.25 & -- \\
\rowcolor{basegrey} & EviViT & 11.19 & 0.745 & 0.841 & 1.091 & 1.187 & 18.17 & \gain{+6.49} \\
\bottomrule
\end{tabular}
\end{table*}

%% file: Tabs/tab_position_controls.tex
\begin{table*}[t]
\centering
\caption{\textbf{Location control with native multi-image input.} Only crop positions change; view count, size, and token budget are fixed. Accuracy (\%); $\Delta$ is the change from the selected locations (pp).}
\label{tab:position_controls}
\tablebodyfont
\setlength{\tabcolsep}{6pt}
\begin{tabular}{lrrrr}
\toprule
Regional locations & VP-All & V$^{\ast}$ & Pooled & $\Delta$ \\
\midrule
\rowcolor{basegrey} Deployed locations & 57.09 & 92.15 & 66.57 & -- \\
Random translation & 47.77 & 82.20 & 57.08 & \loss{-9.49} \\
Cross-image shuffled locations & 47.77 & 81.68 & 56.94 & \loss{-9.63} \\
\bottomrule
\end{tabular}
\end{table*}

%% file: Tabs/tab_paired_audit.tex
\begin{table*}[t]
\centering
\caption{\textbf{Questions rescued and lost after attaching EviViT.} Rescue/loss counts answers that become correct/incorrect. Base, +EviViT and Retain are percentages; $\Delta$ is in pp. Retain counts the share of originally correct answers that stay correct; $p$ is the two-sided exact McNemar $p$-value. Qwen3-VL and Qwen3.5 results come from completed answer-only paired runs using the same executor.}
\label{tab:paired_audit}
\tablebodyfont
\setlength{\tabcolsep}{6pt}
\begin{tabular}{@{}llrrrrrr@{}}
\toprule
Model & VP tier & Base & +EviViT & $\Delta$ & Rescue/loss & Retain & $p$ \\
\midrule
Qwen3-VL-4B & Easy & 60.28 & 63.83 & \gain{+3.55} & 16/11 & 87.06\% & 0.44 \\
 & Medium & 40.30 & 50.37 & \gain{+10.07} & 41/14 & 87.04\% & 3.6e-04 \\
 & Hard & 41.51 & 48.11 & \gain{+6.60} & 17/10 & 77.27\% & 0.25 \\
Qwen3-VL-8B & Easy & 67.38 & 69.50 & \gain{+2.13} & 14/11 & 88.42\% & 0.69 \\
 & Medium & 42.16 & 53.36 & \gain{+11.19} & 45/15 & 86.73\% & 1.3e-04 \\
 & Hard & 44.34 & 51.89 & \gain{+7.55} & 19/11 & 76.60\% & 0.2 \\
Qwen3.5-4B & Easy & 63.83 & 70.92 & \gain{+7.09} & 19/9 & 90.00\% & 0.087 \\
 & Medium & 44.40 & 53.36 & \gain{+8.96} & 36/12 & 89.92\% & 7.2e-04 \\
 & Hard & 45.28 & 55.66 & \gain{+10.38} & 20/9 & 81.25\% & 0.061 \\
Qwen3.5-9B & Easy & 65.96 & 66.67 & \gain{+0.71} & 17/16 & 82.80\% & 1 \\
 & Medium & 44.40 & 54.85 & \gain{+10.45} & 37/9 & 92.44\% & 4.1e-05 \\
 & Hard & 42.45 & 51.89 & \gain{+9.43} & 19/9 & 80.00\% & 0.087 \\
\bottomrule
\end{tabular}
\end{table*}

%% file: Tabs/tab_sft_recovery.tex
\begin{table*}[t]
\centering
\caption{\textbf{Language-side SFT with frozen EviViT on Qwen3-VL-8B.} Both runs use the same examples, order and LoRA settings. We report the end of epoch 3. Avg. equally weights Visual-CoT (VCoT), VP-E/M/H and Zoom. Scores are accuracy (\%); green gains (pp) compare each SFT row to its frozen path. Doc/OCR and Other five report equally weighted means over the two groups of Visual-CoT tasks.}
\label{tab:sft_recovery}
\tablebodyfont
\setlength{\tabcolsep}{2.4pt}
\renewcommand{\arraystretch}{1.04}
\begin{tabular}{llrrrrrrrrl}
\toprule
Model & Trainable & VCoT & Doc/OCR & Other five & VP-E & VP-M & VP-H & Zoom & Avg. & {} \\
\midrule
Base & Frozen & 78.50 & 89.53 & 70.86 & 67.38 & 42.16 & 44.34 & 43.91 & 55.26 & {} \\
\rowcolor{basegrey} \method{} & Frozen & 78.31 & 87.94 & 71.82 & 69.50 & 53.36 & 51.89 & 50.65 & 60.74 & {} \\
\midrule
Base+SFT & LLM LoRA & 79.20 & 89.55 & 72.18 & 70.21 & 44.03 & 46.23 & 42.96 & 56.53 & \gain{(+1.27)} \\
\rowcolor{basegrey} \method{}+SFT & LLM LoRA & 79.18 & 88.63 & 73.10 & 70.92 & 53.73 & 53.77 & 50.30 & 61.58 & \gain{(+0.84)} \\
\bottomrule
\end{tabular}
\end{table*}

%% file: Sec/research_statements.tex
\subsection*{Ethics statement}
The annotators were members of the research team. They consented to recording
their own searches on public benchmark images. Because the team collected
only its own interactions and recruited no outside participants, it did not
seek separate ethics committee review.

\subsection*{Reproducibility statement}
Table~\ref{tab:bridge_configuration} records the training settings and compute
resources. Supplementary Sec.~\ref{app:isolated_cost} specifies the timing
protocol, and Sec.~\ref{app:search_budget} defines the visual budgets for
single-call and multi-call evaluation.

%% file: Tabs/tab_visual_cot_taskwise.tex
\begin{table}[t]
\centering
\caption{\textbf{Visual-CoT on the Qwen3.5 models.} Accuracy (\%) with the same 8B judge on each task. E adds EviViT to the base model. The last row pools all questions.}
\label{tab:visual_cot_taskwise}
\small
\setlength{\tabcolsep}{2pt}
\begin{tabular*}{\columnwidth}{@{\extracolsep{\fill}}lrrrrr@{}}
\toprule
& & \multicolumn{2}{c}{Qwen3.5-4B} & \multicolumn{2}{c}{Qwen3.5-9B}\\
\cmidrule(lr){3-4}\cmidrule(l){5-6}
Task & $N$ & Base & +E & Base & +E\\
\midrule
Flickr30K & 1,546 & 75.42 & 79.30 & 76.84 & 79.24 \\
DocVQA & 888 & 96.06 & 96.06 & 96.96 & 96.73 \\
GQA & 978 & 67.38 & 67.28 & 70.86 & 72.70 \\
InfographicsVQA & 360 & 85.83 & 84.44 & 86.39 & 86.39 \\
OpenImages & 945 & 51.75 & 53.33 & 50.05 & 53.54 \\
TextVQA & 526 & 93.54 & 92.59 & 94.49 & 94.68 \\
VSR & 404 & 73.02 & 74.26 & 76.49 & 78.71 \\
DUDE & 602 & 76.91 & 77.24 & 76.41 & 76.91 \\
SROIE & 686 & 96.36 & 96.21 & 95.92 & 96.06 \\
CUB & 492 & 70.12 & 70.93 & 84.55 & 84.76 \\
\midrule
All questions & 7,427 & 77.18 & 78.17 & 78.98 & 80.34 \\
\bottomrule
\end{tabular*}
\end{table}

%% file: Sec/appendix_evidence_audits.tex
\section{Evidence selection and integration}
\label{app:evidence_audits}

\subsection{Re-reading the source at a fixed visual budget}
\label{app:source_paths}
Both paths use the same boxes, view order and encoder grids on each
question, with at most 4,096 visual tokens across all views. Multi-image takes
each crop from the resized global image and enlarges it to the assigned
grid. EviViT samples that grid directly from the source image and fuses
the resulting features inside the encoder. The two paths therefore differ in
both the source pixels and how the model incorporates them; this comparison
measures those choices together.

\input{Tabs/tab_source_paths}

\begin{samepage}
Across the three benchmarks in Table~\ref{tab:source_paths}, EviViT
gives 23 more correct answers than the multi-image path using thumbnail crops. Of these
net gains, 21 occur on VisualProbe and two on ZoomBench. On V$^\ast$ the
two paths answer the same number correctly.\par
\end{samepage}

\subsection{Why the selected locations matter}
\label{app:position_controls}

Would crops of the same size work equally well elsewhere? We test location
alone by giving Qwen3-VL-8B the global view and crops as separate images
through its native multi-image interface. All three conditions use the
same 706 questions. Within each question, we keep the global view, crop
count and order, and each crop's role, pixel dimensions, output grid and
token allocation fixed. The model weights, prompt, decoding and judge
are identical. Only the crop centers change.

Random translation moves each crop to a sampled valid position in the same
image. Cross-image shuffling borrows ordered centers from a different image
in the same benchmark with the same region count, retaining the current
sample's crop sizes. This preserves positions selected for another
image--question pair but breaks their alignment with the current one. Both
controls use a seed fixed before inference and do not consult question
content, answers, predictions, or target annotations.

The selected locations in Table~\ref{tab:position_controls} outperform
both controls on VisualProbe and V$^{\ast}$. Pooled accuracy falls by
9.49~pp under random translation and 9.63~pp when centers are shuffled.
The crops cover a similar share of the image on average. What matters
is whether that area contains evidence for the current question.

\subsection{Coverage by the regions actually deployed}
\label{app:deployed_coverage}

We measure how much of each annotated target falls inside the selected
crops. The V$^\ast$ localization subset in Supplementary
Sec.~\ref{subsec:ablation} contains 240 target boxes across 132 questions
with one target and 54 with two.

We use the saved integer-pixel crop rectangles, excluding the global view.
For target $T_{ik}$ and the union $U_i$
of local crops in sample $i$, coverage is
$c_{ik}=|T_{ik}\cap U_i|/|T_{ik}|$. We report the mean target coverage in each
sample, its minimum target coverage, and the fraction of samples for which
every target reaches 90\% coverage.

Table~\ref{tab:deployed_coverage} shows that the selected regions concentrate
useful detail into a small part of the image: they occupy 12.39\% of its area
while achieving a mean target-area coverage of 73.69\%. To test whether this comes from
the choice of locations, we translate each complete crop layout to 10,000
random valid positions, preserving its sizes, relative positions, overlap, and
exact union area. We also center the same layout, and separately randomize
individual crops while preserving their sizes. These controls use no target annotations. Shifting all crops together
reduces mean target coverage to 18.47\%, even though the union area is
identical. Centering the same layout also leaves a substantial gap.

In questions with two annotated targets, the context crop can include an
object that the detail crops miss.
Adding it to the decisive views increases the number of questions with
at least 90\% coverage of \emph{every} target from 98 to 120 out of 186.
Six of those 22 additional questions require two targets. In five of the
six, the decisive views already cover one target but miss the other
entirely; the context view brings the second target into the crop set.

\input{Tabs/tab_deployed_coverage}

%% file: Tabs/tab_source_paths.tex
\begin{table}[!htb]
\centering
\caption{\textbf{Crop source and input path at 4K.} Accuracy (\%) on Qwen3-VL-8B. Multi-image crops the resized global view; EviViT re-reads the original. Both submit the same tokens on every question. The pooled score uses all 1,551 questions.}
\label{tab:source_paths}
\small
\setlength{\tabcolsep}{5pt}
\renewcommand{\arraystretch}{1.10}
\begin{tabular}{@{}lrrrr@{}}
\toprule
Input & VP & V$^\ast$ & Zoom & Pooled\\
\midrule
Multi-image & 44.08 & 91.10 & 48.99 & 52.55\\
EviViT & \textbf{48.16} & 91.10 & \textbf{49.23} & \textbf{54.03}\\
\bottomrule
\end{tabular}
\end{table}

%% file: Tabs/tab_deployed_coverage.tex
\begin{table}[!ht]
\centering
\caption{\textbf{Coverage of annotated targets by deployed local crops.} We evaluate on the V$^\ast$ localization subset. Mean averages target coverage within each sample, then across samples; Min averages each sample's least-covered target. All90 requires every target to reach 90\% coverage. Area is the share of the image covered by the crops. Both shifts are random. All values are percentages.}
\label{tab:deployed_coverage}
\tablebodyfont
\setlength{\tabcolsep}{6pt}
\begin{tabular}{lrrrr}
\toprule
Region layout & Mean & Min & All90 & Area \\
\midrule
\rowcolor{basegrey} EviViT & 73.69 & 68.28 & 64.52 & 12.39 \\
Joint shift & 18.47 & 15.95 & 12.38 & 12.39 \\
Centered & 25.83 & 22.55 & 15.59 & 12.39 \\
Independent shifts & 14.47 & 11.73 & 9.00 & 12.20 \\
\bottomrule
\end{tabular}
\end{table}